\documentclass{article}
\usepackage{algorithm}
\usepackage{algorithmic}
\usepackage{iclr2020_conference,times}

\usepackage{amsmath,amsfonts,bm}

\def\eqref#1{equation~\ref{#1}}
\def\1{\bm{1}}

\DeclareMathAlphabet{\mathsfit}{\encodingdefault}{\sfdefault}{m}{sl}
\SetMathAlphabet{\mathsfit}{bold}{\encodingdefault}{\sfdefault}{bx}{n}

\usepackage{subcaption}
\usepackage{hyperref}
\usepackage{url}
\usepackage{placeins}
\usepackage{wrapfig}
\usepackage{caption}

\usepackage{nicefrac}
\usepackage{microtype}
\usepackage{graphicx}
\usepackage{float}
\usepackage{booktabs} %

\usepackage{hyperref}

\definecolor{mydarkblue}{rgb}{0,0.08,0.45}

\iclrfinalcopy
\hypersetup{
    bookmarks=true,         % show bookmarks bar?
    unicode=false,          % non-Latin characters in Acrobat’s bookmarks
    pdftoolbar=true,        % show Acrobat’s toolbar?
    pdfmenubar=true,        % show Acrobat’s menu?
    pdffitwindow=false,     % window fit to page when opened
    pdfstartview={FitH},    % fits the width of the page to the window
    pdftitle={My title},    % title
    pdfauthor={Author},     % author
    pdfsubject={Subject},   % subject of the document
    pdfcreator={Creator},   % creator of the document
    pdfproducer={Producer}, % producer of the document
    pdfkeywords={keyword1, key2, key3}, % list of keywords
    pdfnewwindow=true,      % links in new PDF window
    colorlinks=true,       % false: boxed links; true: colored links
    linkcolor=mydarkblue,          % color of internal links (change box color with linkbordercolor)
    citecolor=mydarkblue,        % color of links to bibliography
    filecolor=mydarkblue,      % color of file links
    urlcolor=mydarkblue           % color of external links
}

\title{IMPACT: Importance Weighted Asynchronous Architectures with Clipped Target Networks}
 
\author{Michael Luo \\
UC Berkeley \\
\texttt{michael.luo@berkeley.edu} \\
\And Jiahao Yao \\
UC Berkeley \\
\texttt{jiahaoyao@berkeley.edu}
\AND Richard Liaw \\
UC Berkeley \\
\And Eric Liang \\
UC Berkeley \\
\And Ion Stoica \\
UC Berkeley \\}
\def \agentname {IMPACT}

\def \worker { \pi_{\text{worker}_i} }
\def \target { \pi_{\text{target}} }

\begin{document}

\maketitle
\begin{abstract}
The practical usage of reinforcement learning agents is often bottlenecked by the duration of training time. To accelerate training, practitioners often turn to distributed reinforcement learning architectures to parallelize and accelerate the training process. However, modern methods for scalable reinforcement learning (RL) often tradeoff between the throughput of samples that an RL agent can learn from (sample throughput) and the quality of learning from each sample (sample efficiency). In these scalable RL architectures, as one increases sample throughput (i.e. increasing parallelization in IMPALA \citep{espeholt2018impala}), sample efficiency drops significantly. To address this, we propose a new distributed reinforcement learning algorithm, \agentname.
\agentname{} extends IMPALA with three changes: a target network for stabilizing the surrogate objective, a circular buffer, and truncated importance sampling. In discrete action-space environments, we show that \agentname{} attains higher reward  and, simultaneously, achieves up to 30\% decrease in training wall-time than that of IMPALA. For continuous control environments, \agentname{} trains faster than existing scalable agents while \textit{preserving the sample efficiency of synchronous PPO}.
\end{abstract}

\section{Introduction}

%Efficient Fast and efficient training 

%.

%On the other hands, minimizing the number of samples we needed is another critical issue here, let alone getting every single sample can be extremely costly in the real life. 

%Policy gradient is one of model-free algorithm in the reinforcement learning, and it employ the REINFORCE (\cite{sutton2000policy}) algorithm. it is a simple algorithm but suffers from the high variance for the estimation of the derivative of the expected total rewards. Thus, in practice, people use the large batch size for the policy gradient algorithm. Vanilla policy gradient is not sample efficient. Different variant comes up to boost sample efficiency of the policy gradient. 

% stops here 

%Two trends can be observed for faster training. First, great leaps have been made to maximize sample efficiency, primarily with APEX, PPO(\cite{schulman2017proximal}), and SAC. Orthogonally, simulation-based reinforcement learning greatly benefits whet scaled to multiple CPUs and GPUs, especially in asynchronous architectures such as A3C and IMPALA.

%Yet, these trends have diverged with inherent tradeoffs between sample efficiency and time efficiency. Our experiments as well as that of DeepMind’s reveal that agents operating under asynchronous distributed architectures are unable to learn in continuous-control environments, even with off-policy agents such as DQNs. Conversely, sample-efficient agents, such as PPO, train 5-10x times longer than IMPALA agents even on the same hardware.

Proximal Policy Optimization \citep{schulman2017proximal} is one of the most sample-efficient on-policy algorithms. However, it relies on a synchronous architecture for collecting experiences, which is closely tied to its trust region optimization objective. Other architectures such as IMPALA can achieve much higher throughputs due to the asynchronous collection of samples from workers. Yet, IMPALA suffers from reduced sample efficiency since it cannot safely take multiple SGD steps per batch as PPO can. The new agent, \textbf{Imp}ortance Weighted \textbf{A}synchronous Architectures with \textbf{C}lipped \textbf{T}arget Networks (\textbf{\agentname}), mitigates this inherent mismatch. Not only is the algorithm highly sample efficient, it can learn quickly, training 30 percent faster than IMPALA. At the same time, we propose a novel method to stabilize agents in distributed asynchronous setups and, through our ablation studies, show how the agent can learn in both a time and sample efficient manner. 

%In fact, our agent can learn Atari Pong in under 2.5 million timesteps in five minutes with just 32 CPUs and 1 Titan XP, significantly outperforming vanilla IMPALA on the same hardware. And for harder continuous environments, \agentname can attain 9500 reward for Half Cheetah in under 3 hours with 16 CPUs and 1 Titan XP, beating other existing scalable implementations.

In our paper, we show that the algorithm \agentname{} realizes greater gains by striking the balance between high sample throughput and sample efficiency. In our experiments, we demonstrate in the experiments that \agentname{} exceeds state-of-the-art agents in training time (with same hardware) while maintaining similar sample efficiency with PPO's. The contributions of this paper are as follows:

\begin{enumerate}
\item We show that when collecting experiences asynchronously, introducing a target network allows for a stabilized surrogate objective and multiple SGD steps per batch (Section 3.1).

\item We show that using a circular buffer for storing asynchronously collected experiences allows for smooth trade-off between real-time performance and sample efficiency (Section 3.2).

\item We show that IMPACT, when evaluated using identical hardware and neural network models, improves both in real-time and timestep efficiency over both synchronous PPO and IMPALA (Section 4).

\end{enumerate}

% https://docs.google.com/drawings/d/1JgVAsEjIoOb1uLeEGy8Ubzb-0PXH_wC7KYTe3M6KN1I/edit
\begin{figure*}[h]
    \centering
    \begin{subfigure}[t]{0.31\textwidth}
        \centering
        \includegraphics[width=\textwidth]{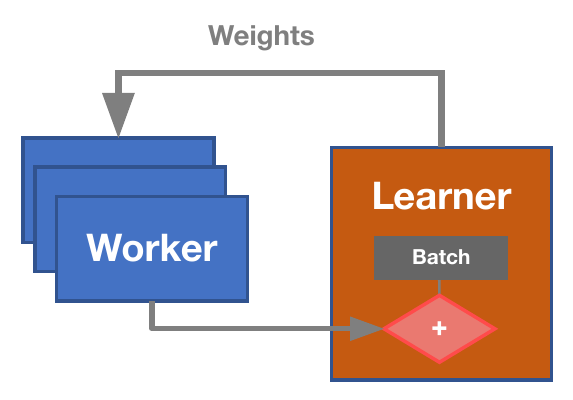}
        \caption{PPO}
        \label{fig:ppo-arch}
    \end{subfigure}
    ~ 
    \begin{subfigure}[t]{0.31\textwidth}
        \centering
        \includegraphics[width=\textwidth]{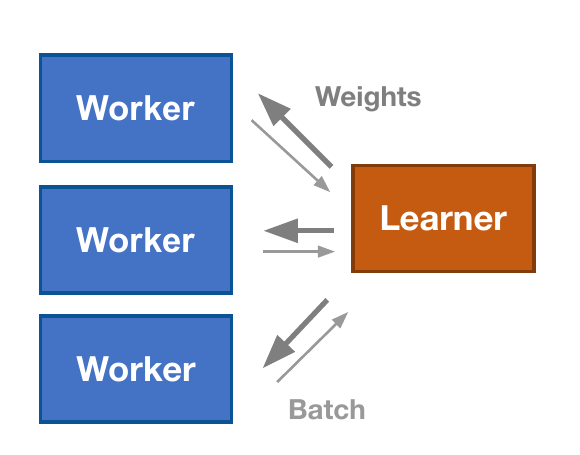}
    \caption{IMPALA}
        \label{fig:impala-arch}
    \end{subfigure}
    ~ 
    \begin{subfigure}[t]{0.31\textwidth}
        \centering
        \includegraphics[width=\textwidth]{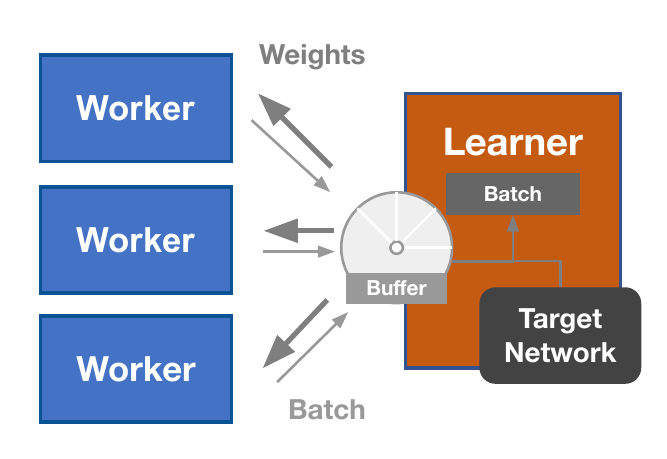}
        \caption{\agentname}
        \label{fig:appo-diag}
    \end{subfigure}
    
    \caption{\small Architecture schemes for  distributed PPO, IMPALA, and \agentname{}. PPO aggregates worker batches into a large training batch and the learner performs minibatch SGD. IMPALA workers asynchronously generate data. \agentname{} consists of a batch buffer that takes in worker experience and a target's evaluation on  the experience. The learner samples from the buffer.}
	\label{fig:rl-arch-diag}
	\vspace{-0.5cm}
 \end{figure*}
\section{Background}
Reinforcement Learning assumes a Markov Decision Process (MDP) setup defined by the tuple $(S,A,p,\gamma,r)$ where $S$ and $A$ represent the state and action space, $\gamma\in [0, 1]$ is the discount factor, and $p:S\times A \times S \rightarrow \mathbb{R}$ and $R: S\times A \rightarrow \mathbb{R}$ are the transition dynamics and reward function that models an environment. 

Let $\pi(a_t|s_t):S \times A \rightarrow [0, 1]$ denote a stochastic policy mapping that returns an action distribution given state $s_t \in S$. Rolling out policy $\pi(a_t|s_t)$ in the environment is equivalent to sampling a trajectory $ \tau \sim \mathbb P( \boldsymbol{ \tau }) $, where $ \tau := (s_0, a_0, ...., a_{T-1}, s_T, a_T) $. We can compactly define state and state-action marginals of the trajectory distribution  $p_\pi(s_t)$ and $p_\pi(s_t,a_t)$ induced by the policy $\pi(a_t|s_t)$.The goal for reinforcement learning aims to maximize the following objective: $ J(\theta) = \mathbb{E}_{(s_t, a_t)\sim p_{\pi}}[\sum_{t=0}^{T} \gamma^t R(s_t,a_t)]$. 

When $\theta$ parameterizes $\pi(a_t|s_t)$, the policy is updated according to the \textbf{Policy Gradient Theorem} \citep{sutton2000policy}:  \[\nabla_{\theta}J(\theta) = \mathbb{E}_{(s_t, a_t) \sim p_\pi(\cdot)}\left[\nabla_{\theta}\log \pi_{\theta}(a_t|s_t)\hat{A}_{\pi_{\theta}}(s_t, a_t) \right],\]
where $ \hat{A}_{\pi_{\theta}}(s_t, a_t) $ is an estimator of the advantage function. The advantage estimator is usually defined as the 1-step TD error, $ \hat{A}_{\pi_{\theta}}(s_t, a_t) = r(s_t, a_t) + \gamma \hat V(s_{t+1}) - \hat V(s_{t})$, where $\hat V(s_{t})$ is an estimation of the value function. Policy gradients, however, suffer from high variance and large update-step sizes, oftentimes leading to sudden drops in performance.  

\subsection{Distributed PPO}
Per iteration, Proximal Policy Optimization (PPO) optimizes policy $\pi_{\theta}$ from target $\pi_{\theta_{\text{old}}}$ via the following objective function
\[L(\theta)=\mathbb{E}_{p_{\pi_{\theta_{\text{old}}}}}\left[\min \left(r_{t}(\theta) \hat{A}_{t}, \operatorname{clip}\left(r_{t}(\theta), 1-\epsilon, 1+\epsilon\right)\hat{A}_{t}\right)\right],\]
where $ r_{t}(\theta) = \frac{\pi_{\theta}(a_t|s_t)}{\pi_{\theta_{\text{old}}}(a_t|s_t)}$ and $\epsilon$ is the clipping hyperparameter. In addition, many PPO implementations use GAE-$\lambda$ as a low bias, low variance advantage estimator for $\hat{A}_t$ \citep{schulman2015high}. PPO's surrogate objective contains the importance sampling ratio $r_{t}(\theta)$, which can potentially explode if $\pi_{\theta_{\text{old}}}$ is too far from $\pi_\theta$. \citep{han2017amber}. PPO's surrogate loss mitigates this with the clipping function, which ensures that the agent makes reasonable steps. Alternatively, PPO can also be seen as an adaptive trust region introduced in TRPO \citep{schulman2015trust}.

In Figure \ref{fig:ppo-arch}, distributed PPO agents implement a synchronous data-gathering scheme. Before data collection, workers are updated to $\pi_{\text{old}}$ and aggregate worker batches to training batch $ D_\text{train}$. The learner performs many mini-batch gradient steps on $ D_\text{train} $. Once the learner is done, learner weights are broadcast to all workers, who start sampling again.

\subsection{Importance Weighted Actor-Learner Architectures}
In Figure \ref{fig:impala-arch}, IMPALA decouples acting and learning by having the learner threads send actions, observations, and values while the master thread computes and applies the gradients from a queue of learner’s experience \citep{espeholt2018impala}. This maximizes GPU utilization and allows for increased sample throughput, leading to high training speeds on easier environments such as Pong. As the number of learners grows, worker policies begin to diverge from the learner policy, resulting in stale policy gradients. To correct this, the IMPALA paper utilizes V-trace to correct the distributional shift: 
\[
v_{s_{t}}=V_{\phi}\left(s_{t}\right)+\sum_{i=t}^{t+n-1} \gamma^{i-t}\left(\prod_{j=t}^{i-1} c_{j}\right) \rho_{i}\left(r_{i+1}+\gamma V_{\phi}\left(s_{i+1}\right)-V_{\phi}\left(s_{i}\right)\right)
\]
where, \(V_{\phi}\) is the value network, \(\pi_{\theta}\) is the policy network of the master thread, \(\mu_{\theta^{\prime}}\) is the policy network of the learner thread, and \(c_{j}=\min\left(\bar{c}, \frac{\pi_{\theta}\left(a_{j} | s_{j}\right)}{\mu_{\theta^{\prime}}\left(a_{j} | s_{j}\right)}\right)\) and \(\rho_{i}=\min\left(\bar{\rho},\frac{\pi_{\theta}\left(a_{i} | s_{i}\right)}{\mu_{\theta^{\prime}}\left(a_{i} | s_{i}\right)}\right)\) are clipped IS ratios.

%\subsection{Wasserstein-2 Distance}
%The Wasserstein distance provides a natural concept of dissimilarity for probability measure. The Wasserstein distance come from the optimal mass transport (OMT) theory (\cite{villani2008optimal}). Wasserstein distance has already achieved success in machine learning, such as Wasserstein GAN (\cite{arjovsky2017wasserstein}).

%Here is the Wasserstein-2 Distance between two probability measure $\mathbb{P}_{r}$ and $\mathbb{P}_{g}$:

%$$
%W_2^2\left(\mathbb{P}_{r}, \mathbb{P}_{g}\right)=\inf _{\gamma \in \Pi\left(\mathbb{P}_{r}, \mathbb{P}_{g}\right)} \mathbb{E}_{(x, y) \sim \gamma}[\|x-y\|^2]
%$$where \(\Pi\left(\mathbb{P}_{r}, \mathbb{P}_{g}\right)\) denotes the set of all joint distributions \(\gamma(x, y)\) whose marginal are \(\mathbb{P}_{r}\) and \(\mathbb{P}_{g} \) respectively. 

\begin{algorithm}[H]
  \caption{\agentname}
  \label{alg:algorithm1}
  \begin{algorithmic}[1]
\REQUIRE Batch size $M$, number of workers $W$, circular buffer size $N$, replay coefficient $K$, target update frequency $t_{\text{target}}$, weight broadcast frequency $t_{\text{frequency}}$, learning rates $\alpha$ and $\beta$\\ 
  \STATE Randomly initialize network weights $(\theta, w)$ \\
  \STATE Initialize target network $(\theta',w')\leftarrow(\theta,w)$ \\
  \STATE Create $W$ workers and duplicate $(\theta,w)$ to each worker
  \STATE Initialize circular buffer $C(N,K)$\\ 
  \FOR {$t=1,..,T$}
      \STATE Obtain batch $B$ of size $M$ traversed $k$ times from $C(N,K)$\\
      \STATE If $k=0$, evaluate $B$ on target $\theta'$, append target output to $B$
      \STATE Compute policy and value network gradients
      \[ \nabla_{\theta}J(\theta)= \frac{1}{M}\sum_{(i,j) \in B } \frac{\nabla_\theta \pi_\theta(s_j | a_j)}{\max(\pi_{\text{target}}(s_j | a_j) , \beta \pi_{{\text{worker}_i}}(s_j| a_j) ) } \hat{A}_{V\text{-GAE}} -\eta \nabla_\theta\mathrm{KL}\left(\pi_{ \text {target } }, \pi_{\theta} \right) \] 
      \[ \nabla_{w}L(w)= \frac{1}{M}\sum_{j}(V_w(s_j) - \hat{V}_{V\text{-GAE}}(s_j))\nabla_{w}V_w(s_j)\]
      \STATE Update policy and value network weights $\theta\leftarrow\theta + \alpha_{t}\nabla_{\theta}J(\theta)$,$w\leftarrow w - \beta_{t}\nabla_{w}L(w)$
      \STATE If $k=K$, discard batch $B$ from $C(N,K)$
      \STATE If $t \equiv 0 \pmod{t_\text{target}}$, update target network $(\theta',w')\leftarrow(\theta,w)$
      \STATE If $t \equiv 0 \pmod{t_\text{frequency}}$, broadcast weights to workers\\
  \ENDFOR
  \end{algorithmic}
  \vspace{0.5em}
\hrule
\vspace{0.3em}
{\bf Worker-i}
\vspace{0.3em}
\hrule
\vspace{0.1em}
\begin{algorithmic}[1]
    \REQUIRE Worker sample batch size $S$
    \REPEAT
        \STATE $B_i = \emptyset$
        \FOR {$t=1, ...,S$}
        \STATE Store $(s_t,a_t,r_t,s_{t+1})$ ran by $\theta_i$ in batch $B_i$
        \ENDFOR
    \STATE Send $B_i$ to $C(N,K)$\\
    \STATE If broadcasted weights exist, set $\theta_i \leftarrow \theta $
    \UNTIL{learner finishes}
\end{algorithmic}

\end{algorithm}

\section{\agentname{} Algorithm}

% https://docs.google.com/drawings/d/1iVNoZjogRDmow7Zjgd-QfqemSEd5pIuMDHu6H193SsE/edit?usp=sharing

% \begin{table*}
%   \begin{tabular}{|c|c|c|c|c|}
%     \hline
% 	&
% 	\textbf{PPO} &
%       \multicolumn{3}{c|}{\textbf{Asynchronous PPO}} \\
%       \hline
%       	\multicolumn{1}{|p{2cm}|}{ \textbf{Invariants} }&$\pi_{\text{worker }}=\pi_{\text{master }}$ &
%       \multicolumn{3}{c|}{ async sampling means \(\pi_{\text {worker }}\) is out of sync with \(\pi_{\text {master }}\)} \\
%       \hline
%       \multicolumn{1}{|p{2cm}|}{ \textbf{Likelihood ratio} }& $\pi_{\theta} / \pi_{\text {worker }} $ & $\pi_{\theta} / \pi_{\text {worker }} $ & $\pi_{\theta} / \pi_{\text {master }}$ & $\min\left( \pi_{\theta} / \pi_{\text {worker }}, \pi_{\theta} / \pi_{\text {target }} \right)$\\
%       \hline
%       \multicolumn{1}{|p{2.25cm}|}{ \textbf{Effectiveness}} & \multicolumn{1}{p{2.6cm}|}{In synchronous
% PPO, all rollouts are
% fully on-policy,
% hence \(\pi_{\text {worker }}\) is the
% same as \(\pi_{\text {master }}\) } &  \multicolumn{1}{p{2.6cm}|}{Since \(\pi_{\text{worker}}\)  may differ
% per worker, using this
% ratio results in trust
% region conflicts across
% multiple batches} & \multicolumn{1}{p{3cm}|}{Since \(\pi_{\text {master }}\) is updated
% after each batch from the
% worker, only a single
% SGD step can be taken
% per batch} & \multicolumn{1}{p{4cm}|}{The IMPACT objective
% allows for multiple SGD
% steps per async batch and
% has a stable trust region.} \\
% \hline
     
%   \end{tabular}
% \end{table*}

\begin{figure*}[h]
    \centering

    \includegraphics[width=\textwidth]{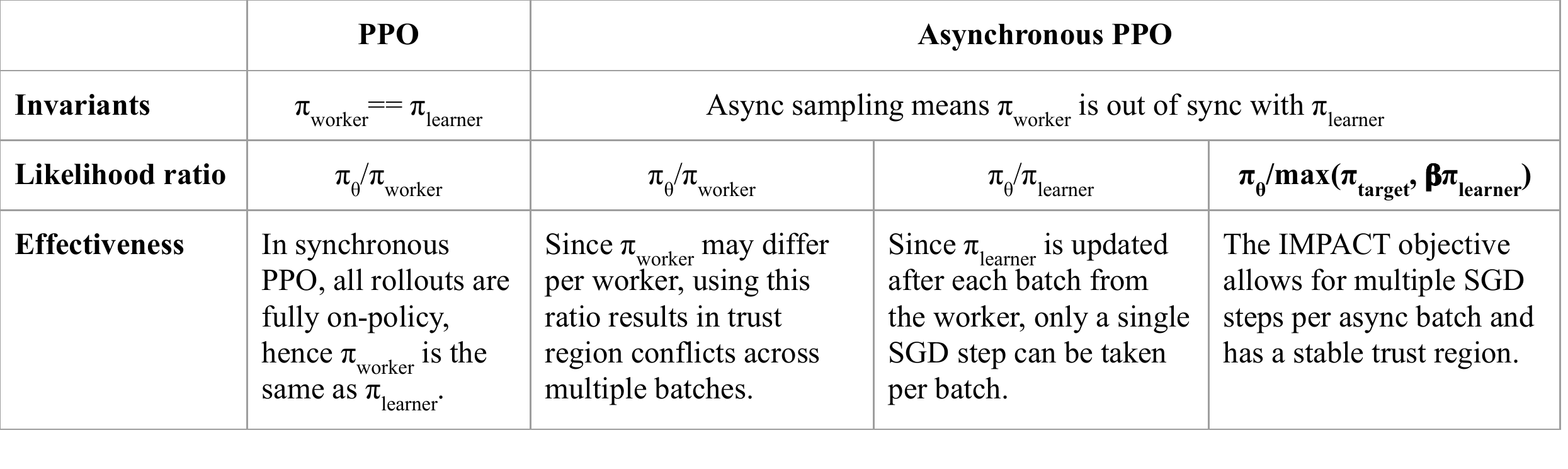}
    \caption{ In asynchronous PPO, there are multiple candidate policies from which the trust region can be defined: (1) $\worker$, the policy of the worker process that produced the batch of experiences, (2) $\pi_{\text{learner}}$, the current policy of the learner process, and (3) $\pi_{\text{target}}$, the policy of a target network. Introducing the target network allows for both a stable trust region and multiple SGD steps per batch of experience collected asynchronously from workers, improving sample efficiency. Since workers can generate experiences asynchronously from their copy of the master policy, this also allows for good real-time efficiency.}
	\label{fig:objective}

\end{figure*}
Like IMPALA, IMPACT separates sampling workers from learner workers. Algorithm~\ref{alg:algorithm1} and Figure~\ref{fig:appo-diag} describe the main training loop and architecture of IMPACT. In the beginning, each worker copies weights from the master network. Then, each worker uses their own policy to collect trajectories and sends the data $(s_t, a_t, r_t)$ to the circular buffer. Simultaneously, workers also asynchronously pull policy weights from the master learner. In the meantime, the target network occasionally syncs with the master learner every $t_{target}$ iterations. The master learner then repeatedly draws experience from the circular buffer. Each sample is weighted by the importance ratio of $\pi_\theta \over \worker$ as well as clipped with target network ratio $\worker \over \pi_{\text{target}}$. The target network is used to provide a stable trust region (Figure \ref{fig:objective}), allowing multiple steps per batch (i.e., like PPO) even in the asynchronous setting (i.e., with the IMPALA architecture). In the next section, we describe the design of this improved objective.

\subsection{Maximal Target-Worker Clipping}

PPO gathers experience from previous iteration's policy $ \pi_{\theta_{\text{old}}}$, and the current policy trains by importance sampling off-policy experience with respect to $\pi_\theta$. 
In the asynchronous setting, worker $i$'s policy, denoted as $\worker$, generates experience for the policy network $ \pi_\theta $. The probability that batch $B$ comes from worker $i$ can be parameterized as a categorical distribution $ i \sim D(\alpha_1, ..., \alpha_n)$. We include this by adding an extra expectation to the importance-sampled policy gradient objective (IS-PG) \citep{jie2010connection}:
\[
J_{IS}(\theta)=\mathbb E_{i \sim D(\alpha)}\left[ \mathbb{E}_{(s_t, a_t)\sim \worker}  \left[\frac{\pi_{\theta}}{\worker}\hat{A}_t\right] \right].
\]

\begin{figure*}[t]
    \centering
    \begin{subfigure}[t]{0.42\textwidth}
        \centering
        \includegraphics[width=\textwidth]{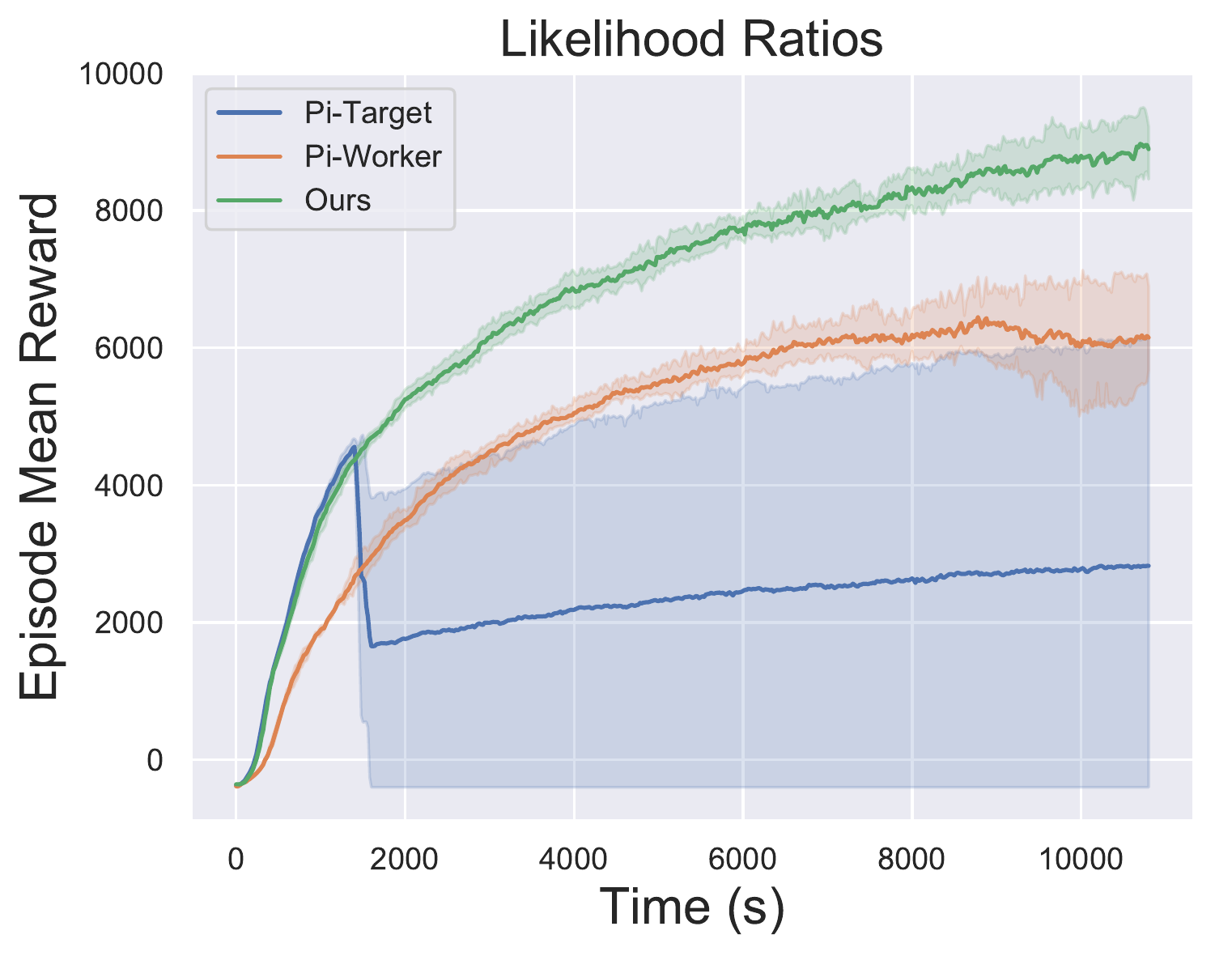}
    \caption{Ratio ablation study.}
        \label{fig:ratio-ablation}
    \end{subfigure}
    % \hspace{20mm}
    \begin{subfigure}[t]{0.46\textwidth}
        \centering
        \includegraphics[width=\textwidth]{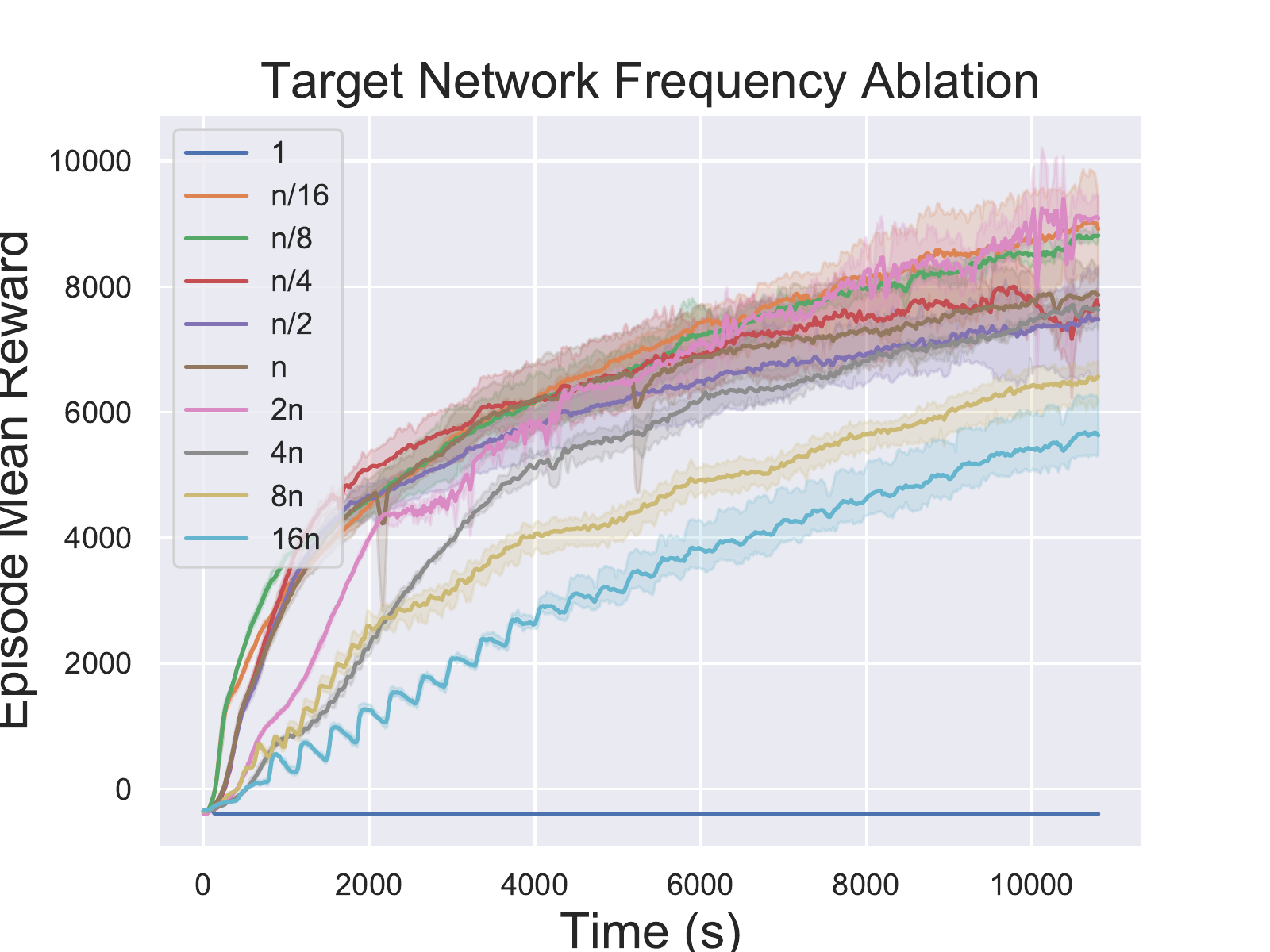}
    \caption{Target update frequency study.}
        \label{fig:ratio-metric-ablation}
    \end{subfigure}
     \caption{\small Training curves of the ablation study on control benchmarks. In (a), the IMPACT objective outperforms other possible ratio choices for the surrogate loss: $
     R_1 = \frac{\pi_\theta}{\pi_{\text{target}}}$, $R_2 = \frac{\pi_\theta}{\pi_{\text{worker}_i}}$, $R_3 = \frac{\pi_\theta}{\max(\pi_{\text{target}} , \beta \worker)}$. In (b), we show the target network update frequency is robust to a range of choices. We try target network update frequency $t_{target}$ equal to the multiple (ranging from 1/16 and 16) of $n=N\cdot K$, the product of the size of circular buffer and the replay times for each batch in the buffer. }
	\label{fig:ablation_curves}
	\hspace{-2mm}
	\vspace{-7mm}
 \end{figure*}

Since each worker contains a different policy, the agent introduces a target network for stability (Figure \ref{fig:objective}). Off-policy agents such as DDPG and DQN update target networks with a moving average. For \agentname, we periodically update the target network with the master network. However, training with importance weighted ratio $\frac{\pi_\theta}{\pi_{\text{target}}}$ can lead to numerical instability, as shown in Figure \ref{fig:ablation_curves}. To prevent this, we clip the importance sampling ratio from worker policy,$\worker$, to target policy, $\pi_{\text{target}}$: 

\begin{align*}
J_{AIS}(\theta)&=\mathbb E_{i \sim D(\alpha)}\left[ \mathbb{E}_{(s_t, a_t)\sim \worker}  \left[\min(\frac{\worker }{ \pi_{\text{target}} }, \rho)\frac{\pi_{\theta}}{\worker}\hat{A}_t\right] \right]\\
&=\mathbb E_{i \sim D(\alpha)}\left[ \mathbb{E}_{(s_t, a_t)\sim \worker}  \left[  \frac{\pi_\theta}{\max(\pi_{\text{target}} , \beta \worker)  }\hat{A}_t\right] \right],
\end{align*}where $\beta=\frac{1}{\rho}$. In the experiments, we set $\rho$ as a hyperparameter with $\rho\geq1$ and $\beta \leq1$. 

To see why clipping is necessary, when master network's action distribution changes significantly over few training iterations, worker i's policy, $\worker$, samples data outside that of target policy, $  \pi_{\text{target}} $, leading to large likelihood ratios, $\frac{\worker}{\pi_{\text{target}}} $. The clipping function $\min(\frac{\worker }{ \pi_{\text{target}} }, \rho)$ pulls back large IS ratios to $\rho$. Figure \ref{fig:intuition curves} in Appendix E provides additional intuition behind the target clipping objective. We show that the target network clipping is a lower bound of the IS-PG objective. 

For $\rho > 1$, the clipped target ratio is larger and serves to augment advantage estimator $\hat{A}_t$. This incentivizes the agent toward good actions while avoiding bad actions. Thus, higher values of $\rho$ encourages the agent to learn faster at the cost of instability. 

% The first term clips the relative ration of $$

% For the likelihood ratio $  \frac{\pi_\theta}{\max(\pi_{\text{target}} , \beta \pi_{\theta_i})  } $, if $ \beta \pi_{\theta_i} > \pi_{\text{target}}$, the ratio becomes the scaled ratio of importance sampling $ \frac{\pi_\theta}{\beta \pi_{\theta_i}  } $; if $ \beta \pi_{\theta_i} < \pi_{\text{target}}$, the ratio becomes $\frac{\pi_\theta}{ \pi_{\text{target}}   }$, which is the lower bounds of the importance sampling.

% We find that this modification allows for successful policy-gradient training in continuous environments, later .

% Experience that workers dynamically generate are run through the target network to generate logits for objective function.

We use GAE-$\lambda$ with V-trace \citep{han2019dimension}. The V-trace GAE-$\lambda$ modifies the advantage function by adding clipped importance sampling terms to the summation of TD errors:
\[
\hat{A}_{V\text{-GAE}}=\sum_{i=t}^{t+n-1} (\lambda\gamma)^{i-t}\left(\prod_{j=t}^{i-1} c_{j}\right)\delta_i V,
\]
where $c_i =\min\left(\bar{c},\frac{\target(a_j|s_j)}{\pi_{\text{worker}_i}(a_j|s_j)}\right)$ (we use the convention $\prod_{j=t}^{t-1} c_{j} = 1$) and $\delta_i V$ is the importance sampled 1-step TD error introduced in V-trace.

\subsection{Circular Buffer}

\begin{figure*}[t]
    \centering
    \begin{subfigure}[t]{0.30\textwidth}
        \centering
        \includegraphics[width=\textwidth]{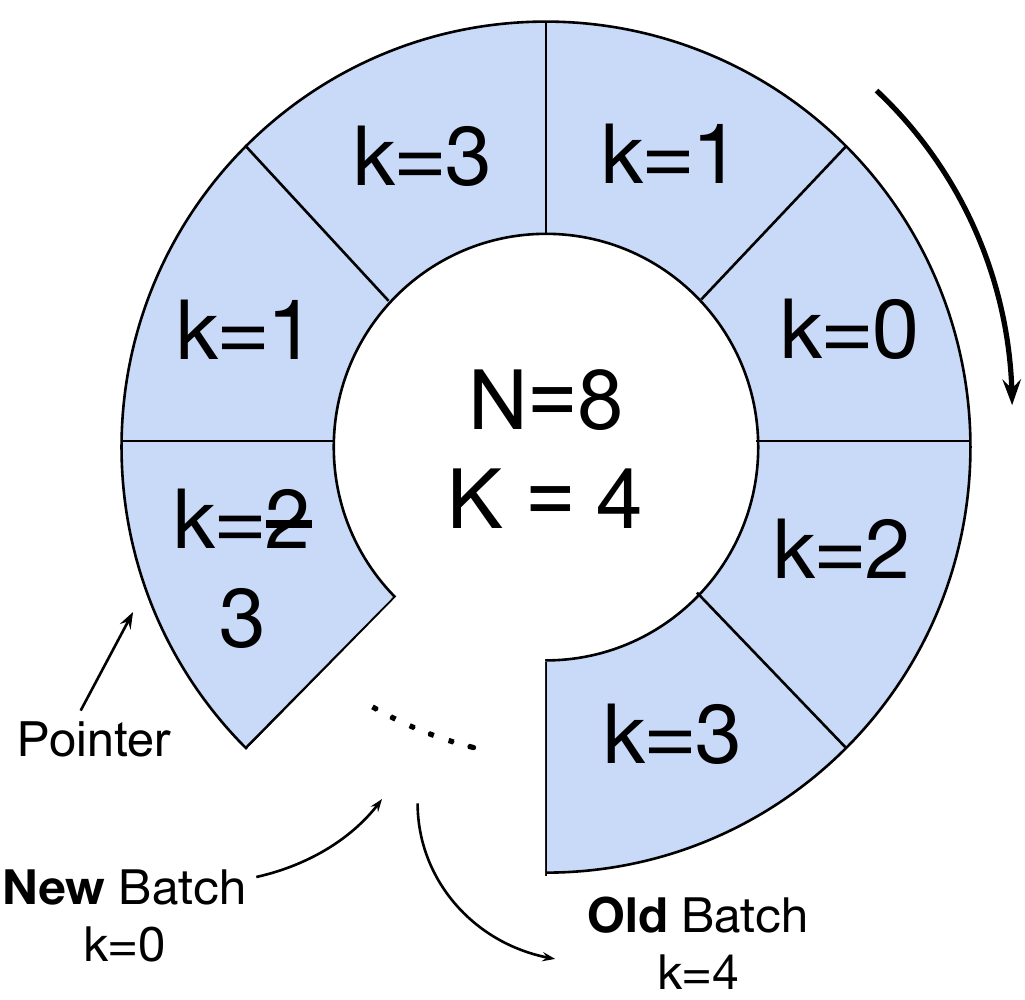}
    \caption{Circular Buffer.}
        \label{fig:circular-buffer-drawing}
    \end{subfigure}
    % \hspace{20mm}
    \begin{subfigure}[t]{0.34\textwidth}
        \centering
        \includegraphics[width=\textwidth]{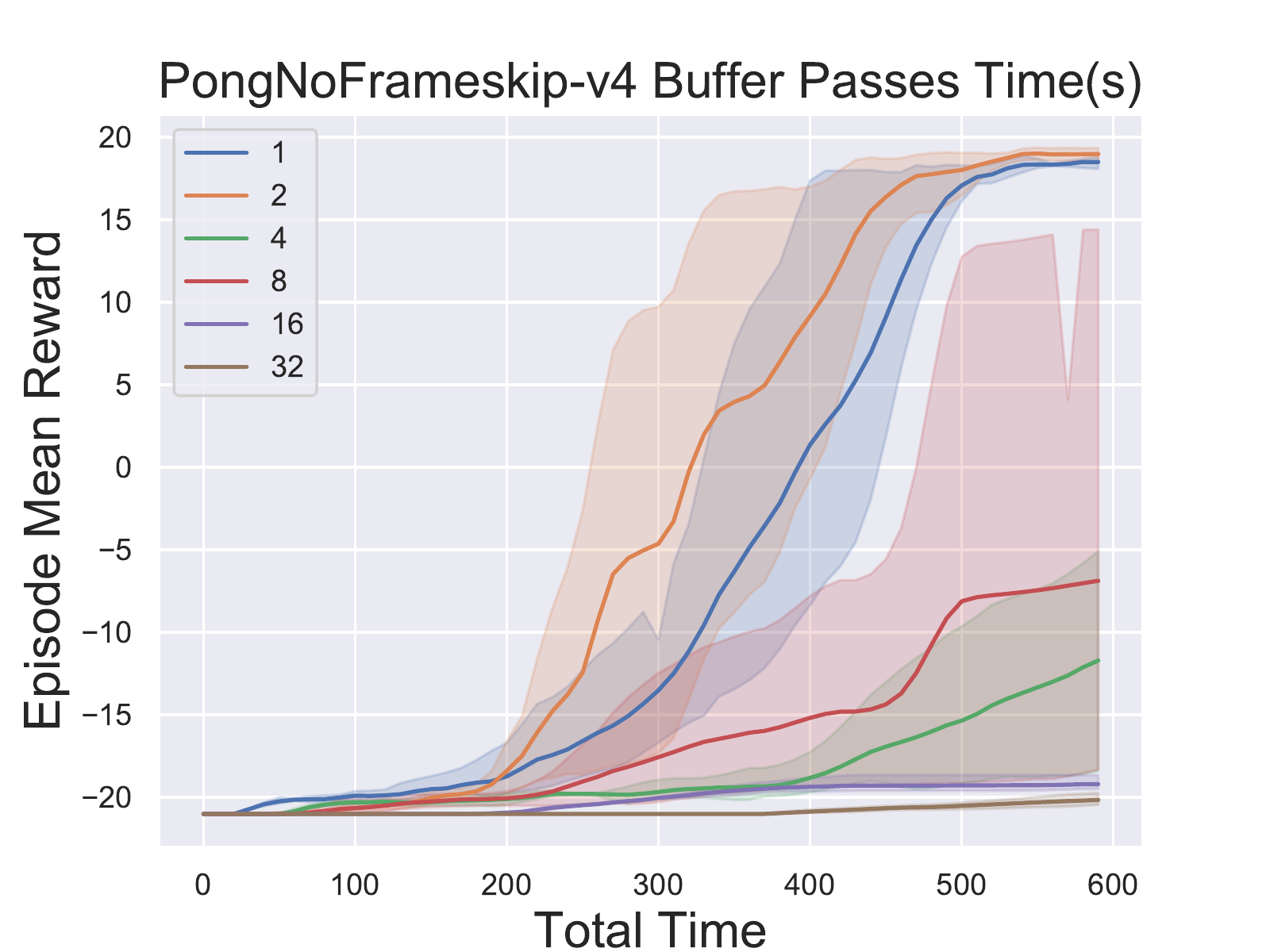}
    \caption{Wall Clock-time vs. $K$}
        \label{fig:circular-buffer-time}
    \end{subfigure}
    \begin{subfigure}[t]{0.34\textwidth}
        \centering
        \includegraphics[width=\textwidth]{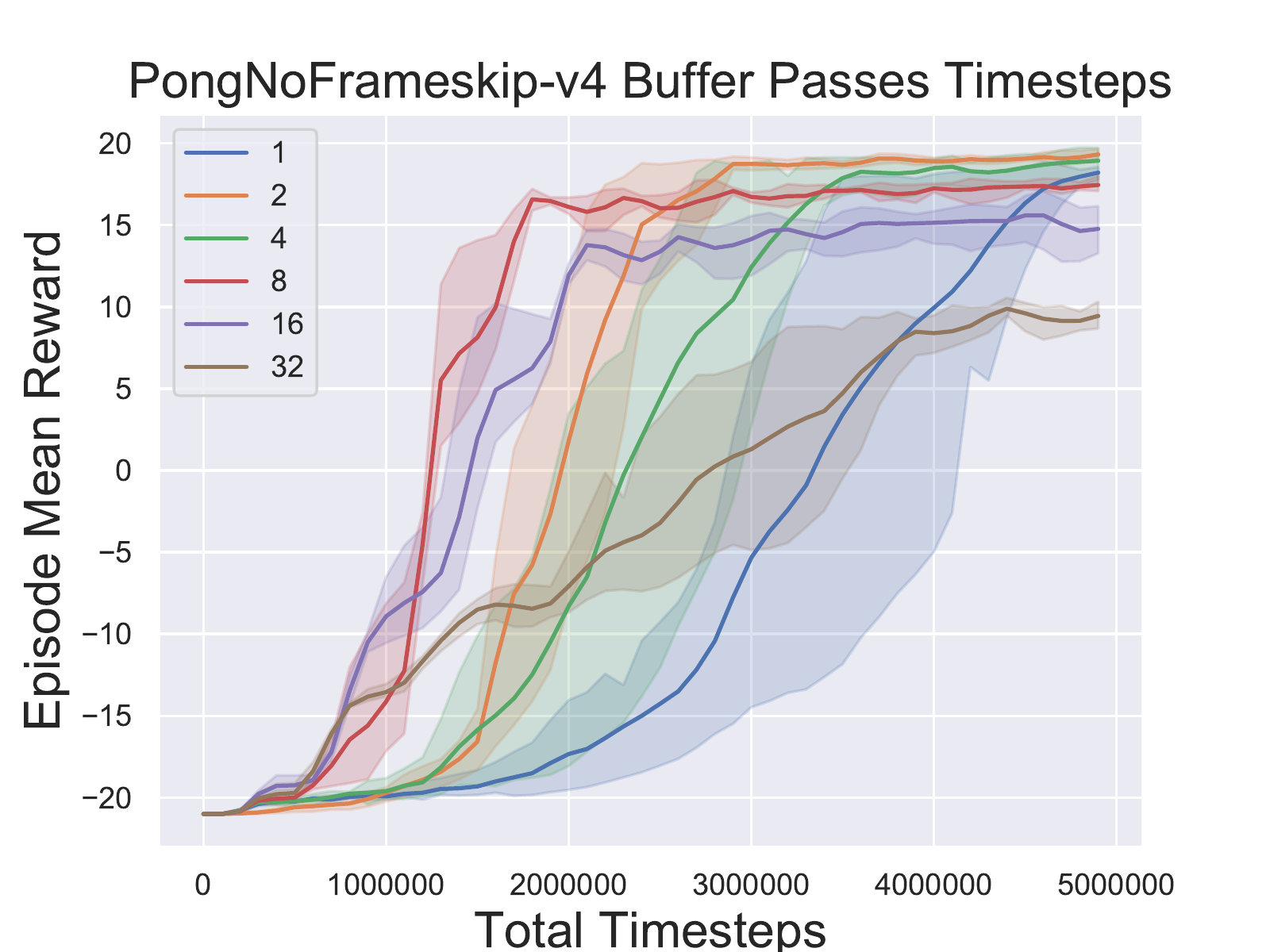}
    \caption{Sample Efficiency vs. $K$}
        \label{fig:circular-buffer-timesteps}
    \end{subfigure}
    \caption{\textbf{(a)}: The Circular Buffer in a nutshell: $N$ and $K$ correspond to buffer size and max times a batch can be traversed. Old batches are replaced by worker-generated batches. \textbf{(b)}: The performance of IMPACT with different K in terms of time.  \textbf{(c)}: The performance of IMPACT with different K in terms of timesteps. IMPACT can achieve greater timestep as well as time efficiency by manipulating $K$. $K=2$ outperforms all other settings in time and is more sample efficient than $K=1,4,16,32$.}
	\label{fig:circularbuffer}
	\hspace{-2mm}
 \end{figure*}

\agentname{} uses a circular buffer (Figure~\ref{fig:circularbuffer}) to emulate the mini-batch SGD used by standard PPO. The circular buffer stores $N$ batches that can be traversed at max $K$ times. Upon being traversed $K$ times, a batch is discarded and replaced by a new worker batch.

For motivation, the circular buffer and the target network are analogous to mini-batching from $\pi_\text{old}$ experience in PPO. When target network's update frequency $n=NK$, the circular buffer is equivalent to distributed PPO's training batch when the learner samples $N$ minibatches for $K$ SGD iterations.

This is in contrast to standard replay buffers, such as in ACER and APE-X, where transitions $(s_t, a_t, r_t, s_{t+1})$ are either uniformly sampled or sampled based on priority, and, when the buffer is full, the oldest transitions are discarded \citep{wang2016sample, horgan2018distributed}.

%Furthermore, on-policy methods can be negatively affected when trained on %off-policy data, and the replay buffer contains a higher portion of %off-policy data. To address this, 
 
Figure~\ref{fig:circularbuffer} illustrates an empirical example where tuning $K$ can increase training sample efficiency and decrease training wall-clock time.

\section{Evaluation}

In our evaluation we seek to answer the following questions:
\begin{enumerate}
 	\item How does the target-clipping objective affect the performance of the agents compared to prior work? (Section 4.1)
 	\item How does the \agentname{} circular buffer affect sample efficiency and training wall-clock time? (Section 4.2)
 	\item How does \agentname{} compare to PPO and IMPALA baselines in terms of sample and real-time performance? (Section 4.3)
 	\item How does \agentname{} scale with respect to the number of workers? (Section 4.4)
\end{enumerate}

\subsection{Target Clipping Performance}

We investigate the performance of the clipped-target objective relative to prior work, which includes PPO and IS-PG based objectives. Specifically, we consider the following ratios below:
$$\begin{array}{ccc}
     R_1 = \frac{\pi_\theta}{\pi_{\text{target}}}  &  R_2 = \frac{\pi_\theta}{\pi_{\text{worker}_i}} &  R_3 = \frac{\pi_\theta}{\max(\pi_{\text{target}} , \beta \worker)}
\end{array}$$

For all three experiments, we truncate all three ratios with PPO's clipping function: $c(R) = \text{clip}(R, 1-\epsilon, 1+\epsilon)$ and train in an asynchronous setting.  Figure 4(a) reveals two important takeaways: first, $R_1$ suffers from sudden drops in performance midway through training. Next, $R_2$ trains stably but does not achieve good performance.

We theorize that $R_1$ fails due to the target and worker network mismatch. During periods of training where the master learner undergoes drastic changes, worker action outputs vastly differ from the learner outputs, resulting in small action probabilities. This creates large ratios in training and destabilizes training.
We hypothesize that $R_2$ fails due to different workers pushing and pulling the learner in multiple directions. The learner moves forward with the most recent worker's suggestions without developing a proper trust region, resulting in many worker's suggestions conflicting with each other. 

The loss function, $R_3$ shows that clipping is necessary and can help facilitate training. By clipping the target-worker ratio, we make sure that the ratio does not explode and destabilize training. Furthermore, we prevent workers from making mutually destructive suggestions by having a target network provide singular guidance.

\subsubsection{Target Network Update Frequency}
In Section 3.2, an analogy was drawn between PPO's mini-batching mechanism and the circular buffer. Our primary benchmark for target update frequency is $ n = N\cdot K$, where $N$ is circular buffer size and $K$ is maximum replay coefficient. This is the case when PPO is equivalent to IMPACT.

In Figure 4(b), we test the frequency of updates with varying orders of magnitudes of $n$. In general, we find that agent performance is robust to vastly differing frequencies. However, when $n=1\sim4$, the agent does not learn. Based on empirical results, we theorize that the agent is able to train as long as a stable trust region can be formed. On the other hand, if update frequency is too low, the agent is stranded for many iterations in the same trust region, which impairs learning speed.

\subsection{Time and Sample Efficiency with Circular Buffer}
Counter to intuition, the tradeoff between time and sample efficiency when $K$ increases is not necessarily true. In Figure \ref{fig:circular-buffer-time} and \ref{fig:circular-buffer-timesteps}, we show that IMPACT realizes greater gains by striking the balance between high sample throughput and sample efficiency. When $K=2$, IMPACT performs the best in both time and sample efficiency. Our results reveal that wall-clock time and sample efficiency can be optimized based on tuning values of $K$ in the circular buffer.

\subsection{Comparison with Baselines}

% \begin{figure*}[!hbtp]
\begin{figure}
    \centering
    \begin{subfigure}{\textwidth}
        \centering
        \includegraphics[width=0.31\textwidth, height=0.15\textheight, clip]{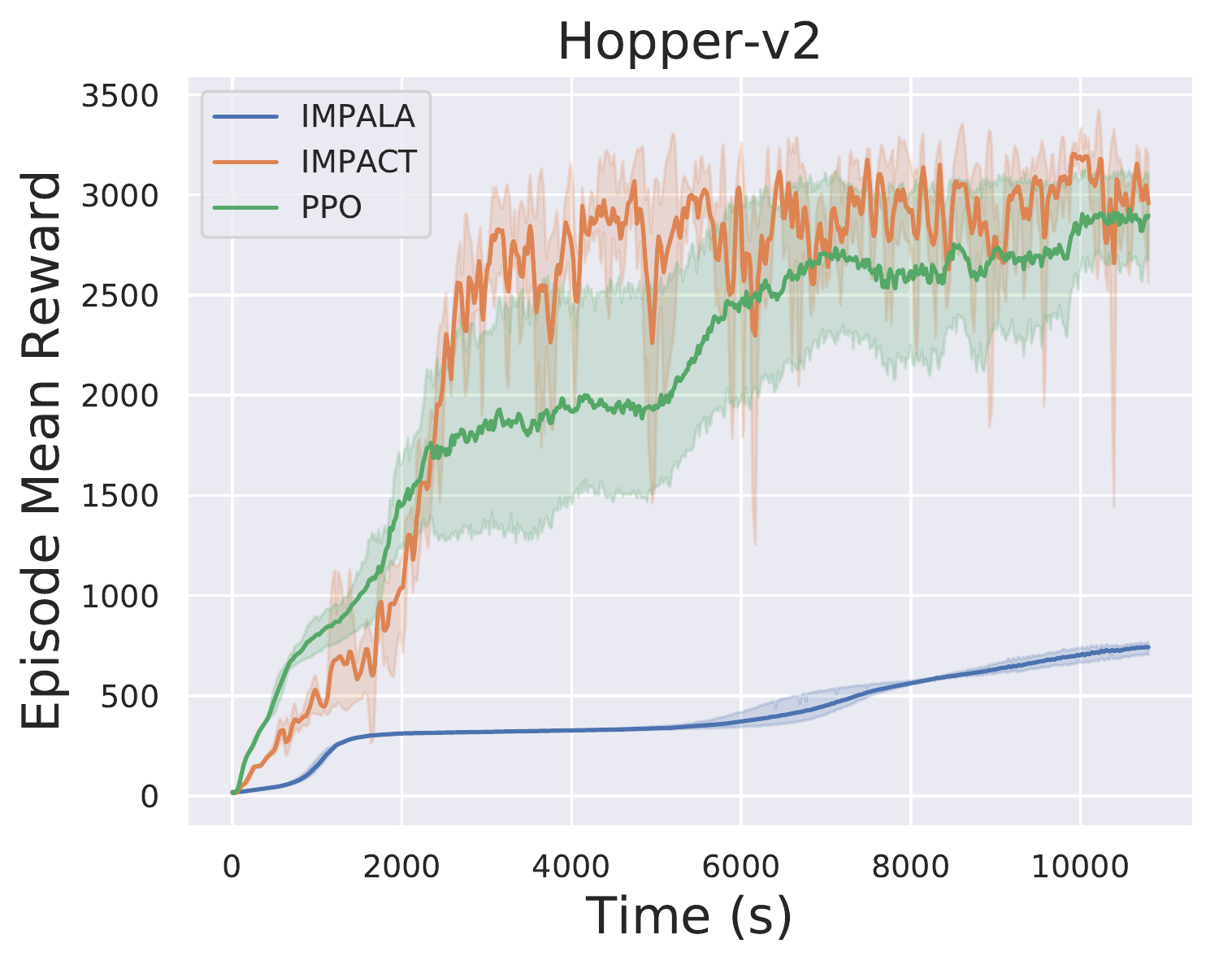}
        \hfill
        \includegraphics[width=0.31\textwidth, height=0.15\textheight, clip]{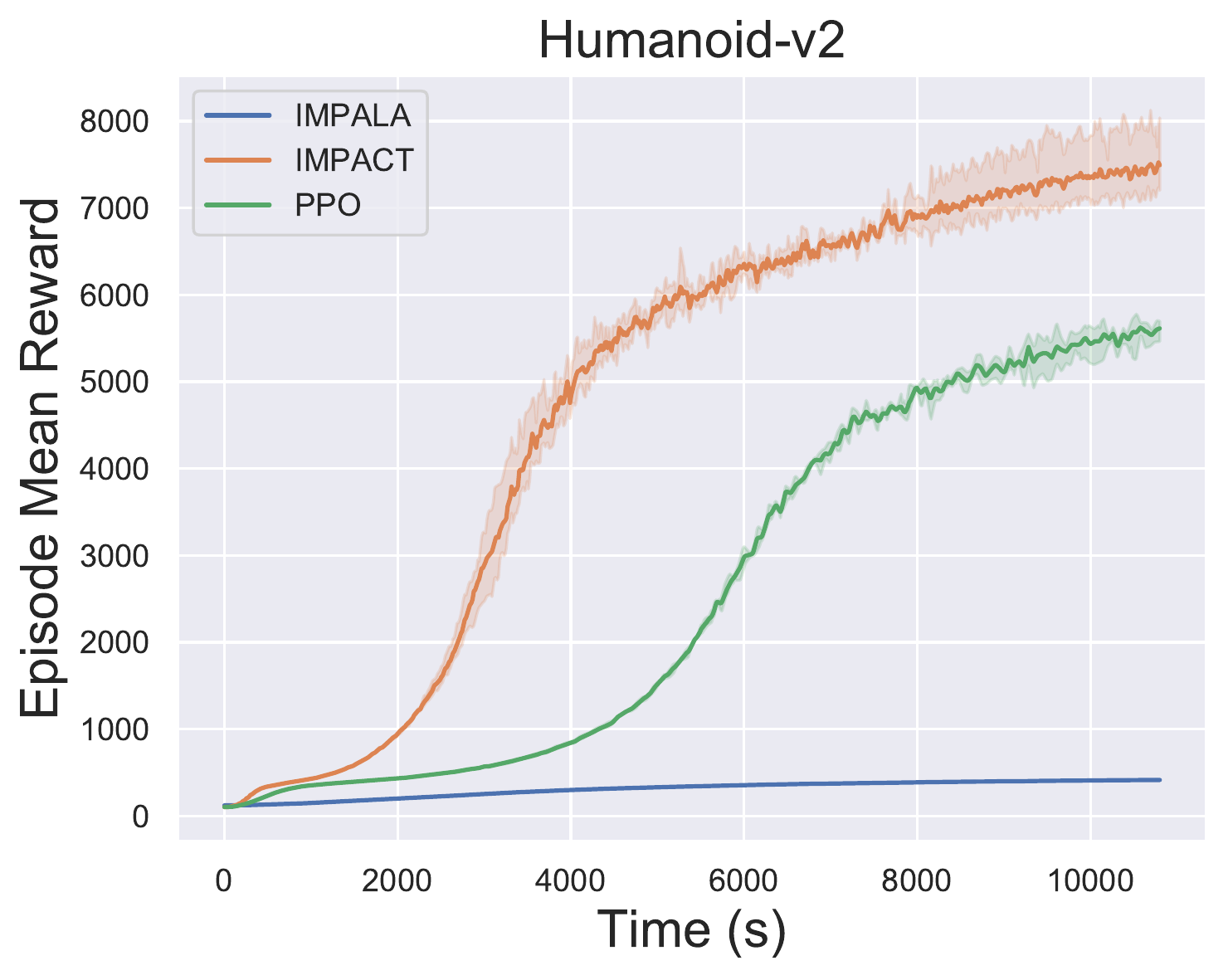}
        \hfill
        \includegraphics[width=0.31\textwidth, height=0.15\textheight, clip]{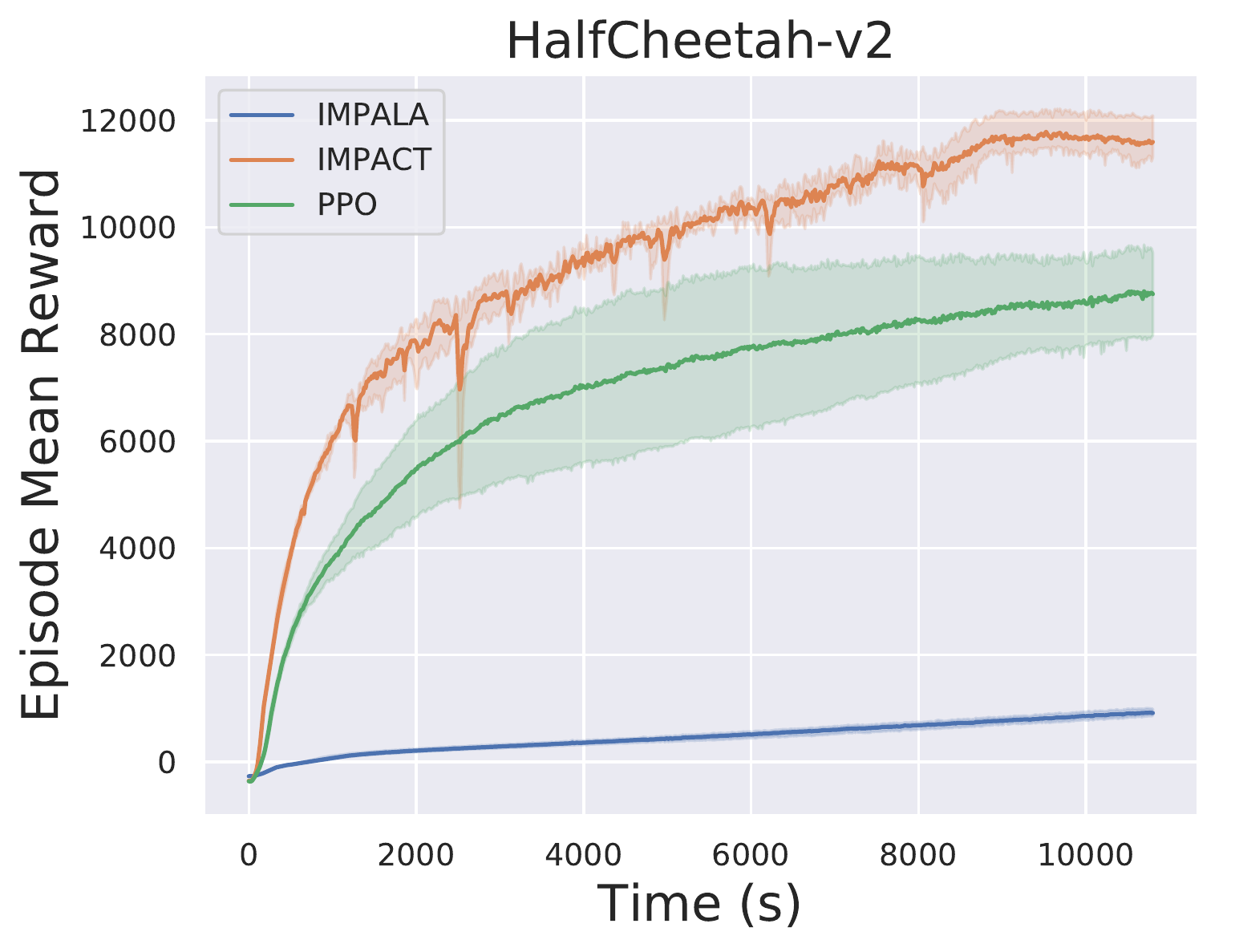}
        \subcaption{\small Time}
    \end{subfigure}
    \vskip\baselineskip 
    \begin{subfigure}{\textwidth}
        \centering
        \includegraphics[width=0.31\textwidth, height=0.15\textheight, clip]{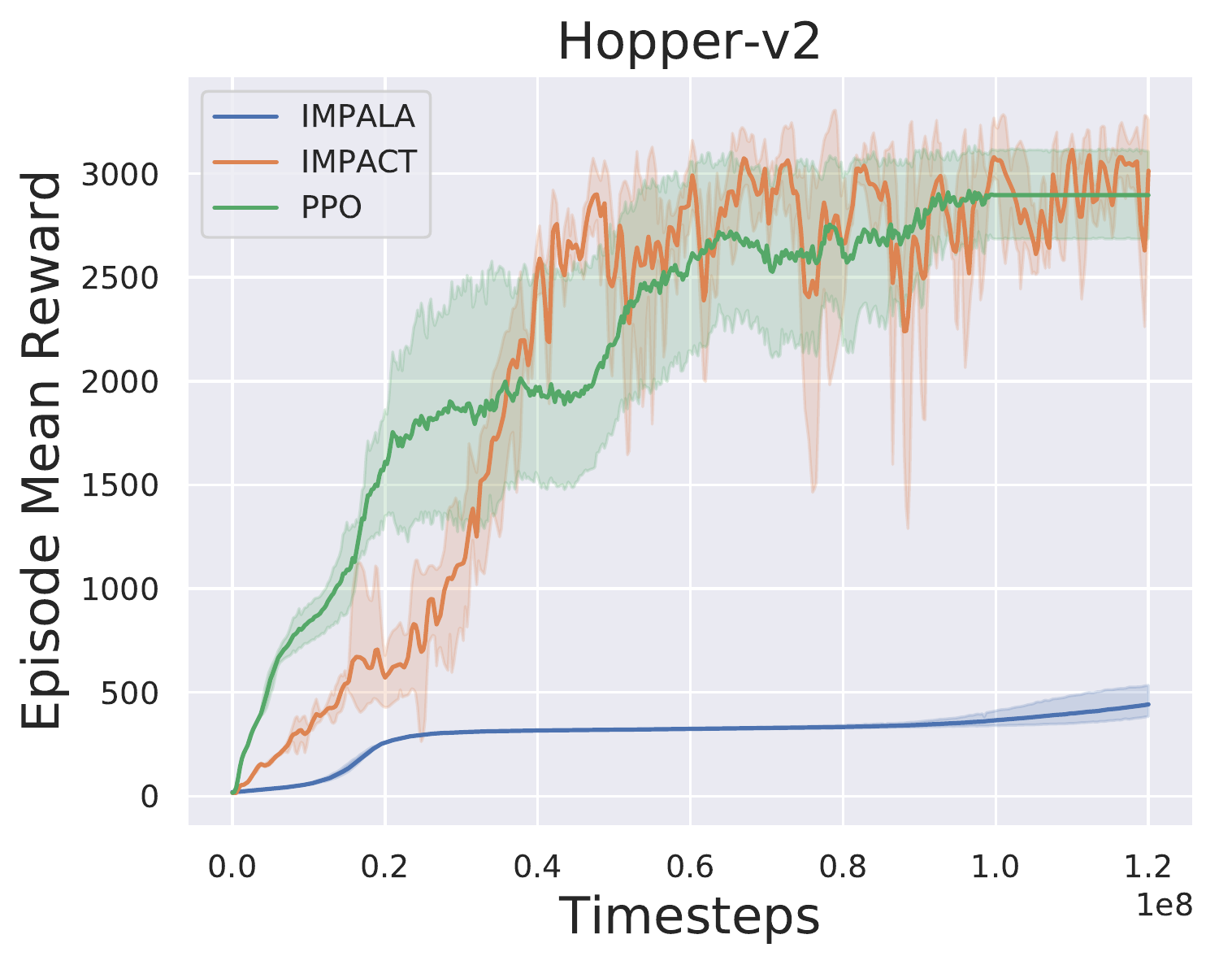}
        \hfill
        \includegraphics[width=0.31\textwidth, height=0.15\textheight, clip]{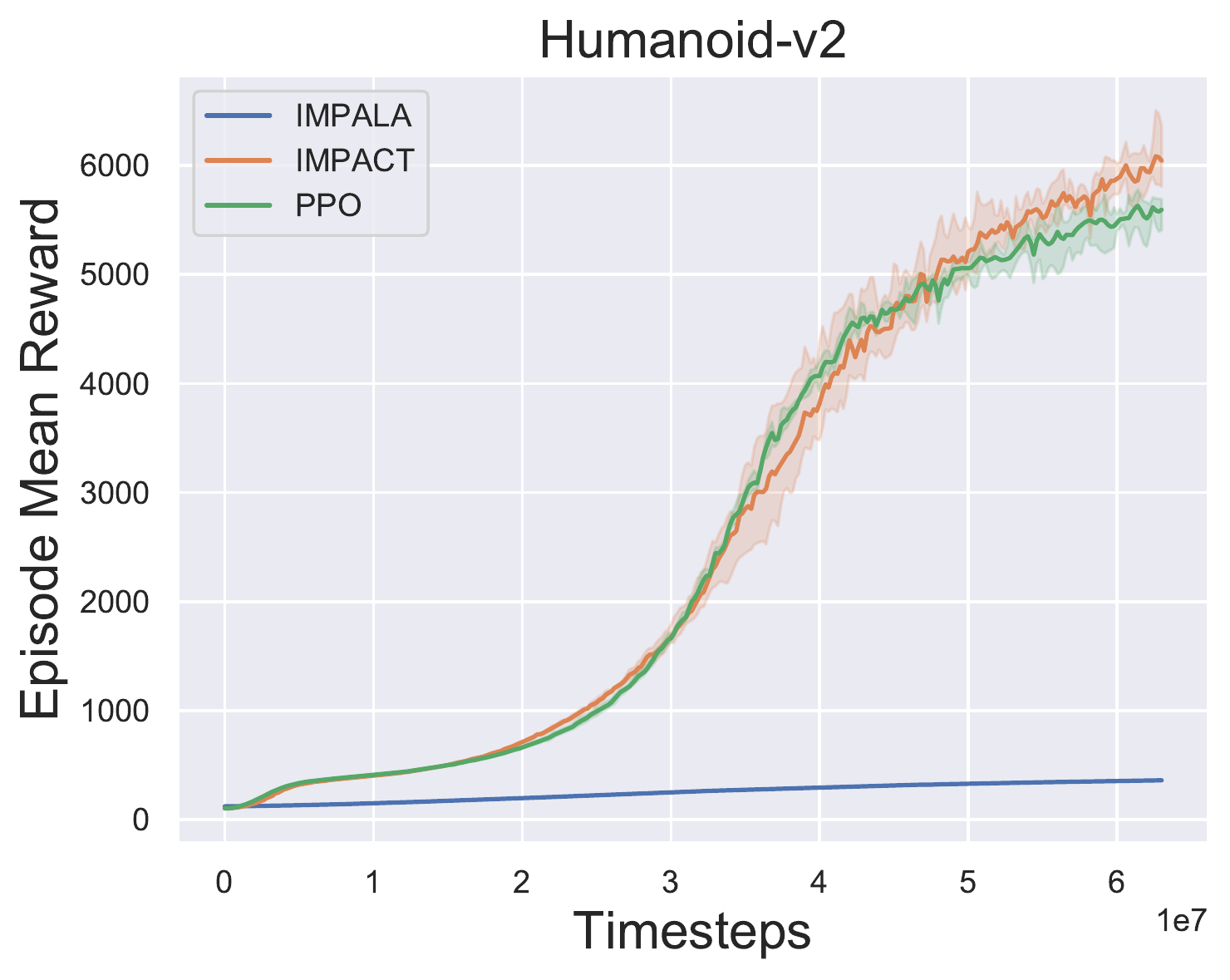}
        \hfill
        \includegraphics[width=0.31\textwidth, height=0.15\textheight, clip]{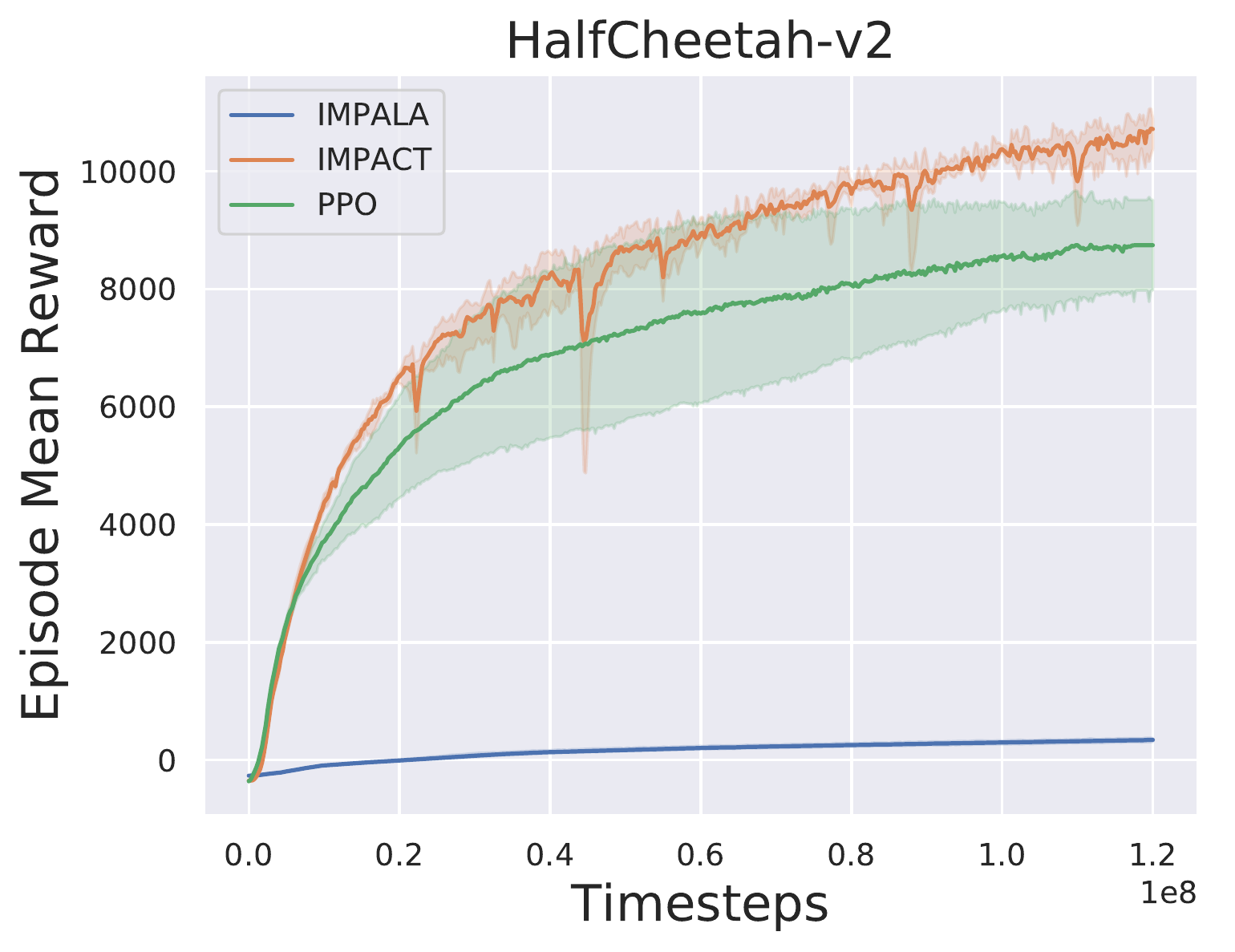}
        \subcaption{\small Timesteps}
    \end{subfigure}
    \caption{\agentname{} outperforms baselines in both sample and time efficiency for Continuous Control Domains: Hopper, Humanoid, HalfCheetah.}
	\label{fig:training_curves}
	\hspace{2mm}
	\vspace{-2mm}
\end{figure}
%  \end{figure*}

We investigate how \agentname{} attains greater performance in wall clock-time and sample efficiency compared with PPO and IMPALA across six different continuous control and discrete action tasks.
 
We tested the agent on three continuous environments (Figure 5): HalfCheetah, Hopper, and Humanoid on 16 CPUs and 1 GPU. The policy networks consist of two fully-connected layers of 256 units with nonlinear activation tanh.  The critic network shares the same architecture as the policy network. For consistentency, same network architectures were employed across PPO, IMPALA, and \agentname.

For the discrete environments (Figure 6), Pong, SpaceInvaders, and Breakout were chosen as common benchmarks used in popular distributed RL libraries \citep{intelcoach, rllib}. Additional experiments for discrete environments are in the Appendix. These experiments were ran on 32 CPUs and 1 GPU. The policy network consists of three 4x4 and one 11x11 conv layer, with nonlinear activation ReLU. The critic network shares weights with the policy network. The input of the network is a stack of four 42x42 down-sampled images of the Atari environment. The hyper-parameters for continuous and discrete environments are listed in the Appendix \ref{sec:exptparams} table \ref{table:discrete_control_settings} and \ref{table:continuous_control_settings} respectively.

% We ran our experiments for three hours for SpaceInvaders and Breakout, but only for .

% In our three, IMPALA was unable to perform well. We tried hyper-parameter turning and finetuning learning rate; however, the asynchronous experience seemed to severely prevent successful learning, even with V-trace. 

Figures 5 and 6 show the total average return on evaluation rollouts for \agentname{}, IMPALA and PPO. We train each algorithm with three different random seeds on each environment for a total time of three hours. According to the experiments, \agentname{} is able to train much faster than PPO and IMPALA in both discrete and continuous domains, while preserving same or better sample efficiency than PPO.

% We have shown that our agent is a stable medium between speed and sample efficiency, achieving greater results.

Our results reveal that continuous control tasks for \agentname{} are sensitive to the tuple $ (N , K) $ for the circular buffer. 
$N=16$ and $K=20$ is a robust choice for continuous control. Although higher $K$ inhibits workers' sample throughput, increased sample efficiency from replaying experiences results in an overall reduction in training wall-clock time and higher reward. For discrete tasks, $N=1$ and $K=2$ works best. Empirically, agents learn faster from new experience than replaying old experience, showing how exploration is crucial to achieving high asymptotic performance in discrete enviornments.

\begin{figure}
    \centering
    \begin{subfigure}{\textwidth}
        \centering
        \includegraphics[width=0.31\textwidth, height=0.15\textheight, clip]{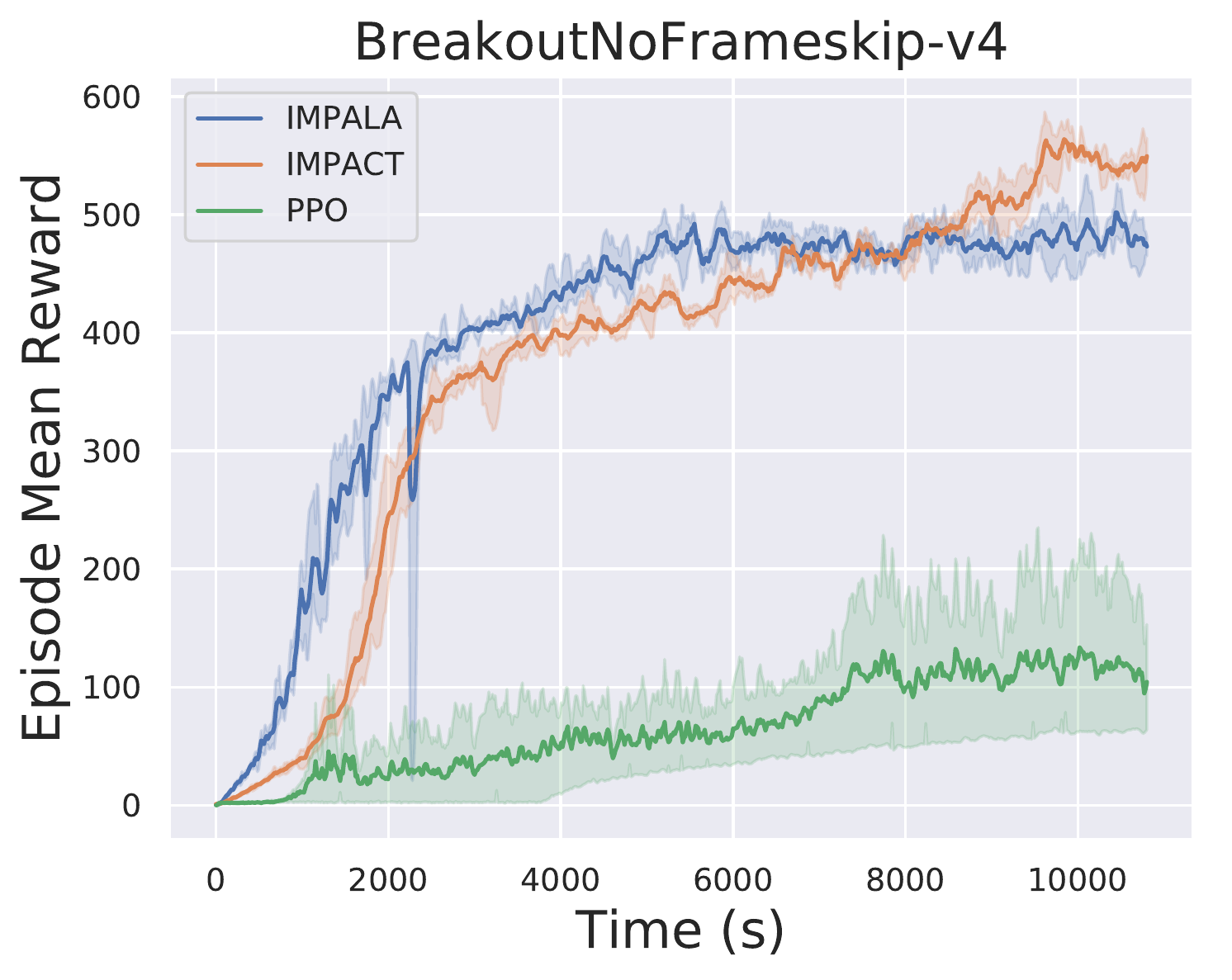}
        \hfill
        \includegraphics[width=0.31\textwidth, height=0.15\textheight, clip]{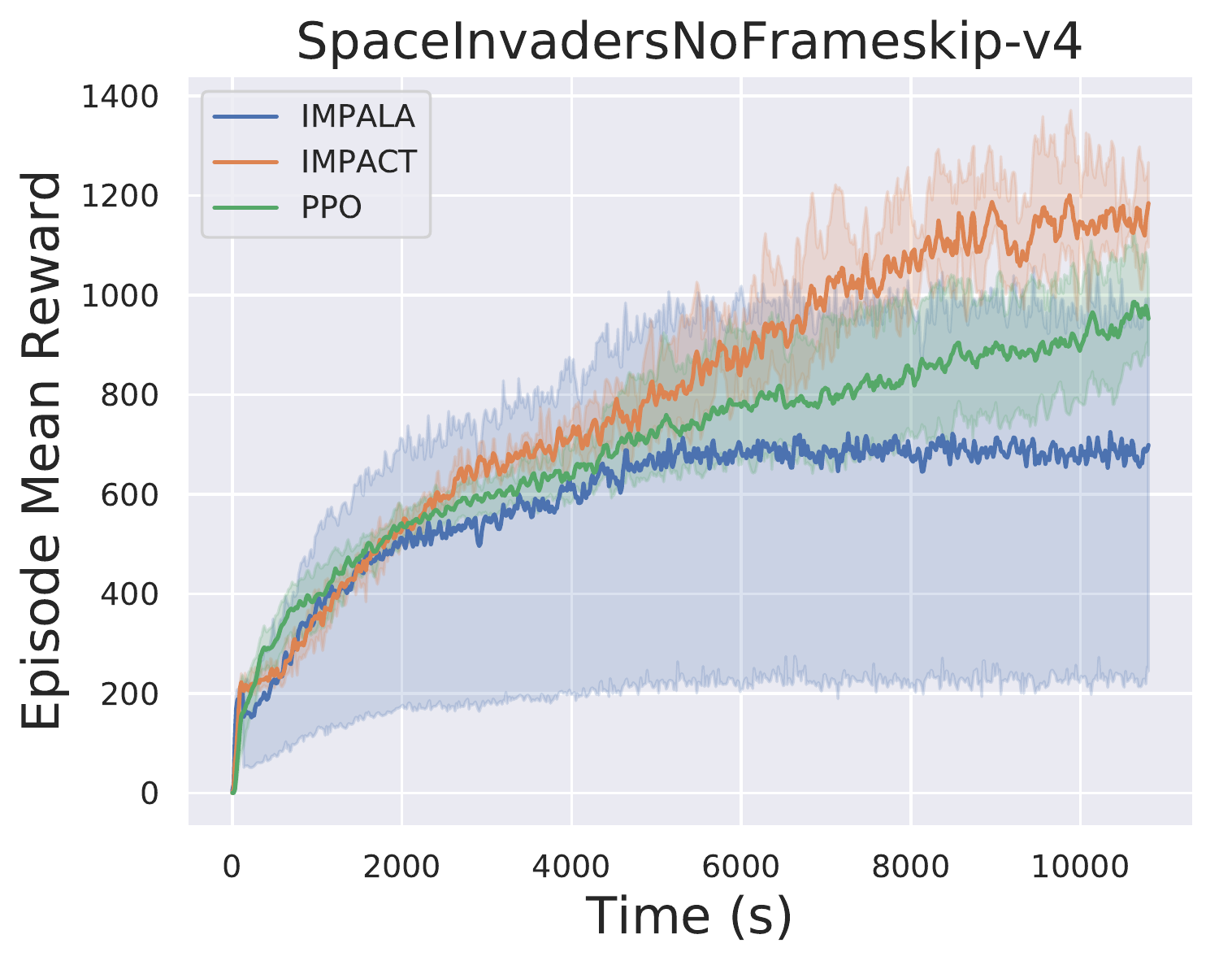}
        \hfill
        \includegraphics[width=0.31\textwidth, height=0.15\textheight, clip]{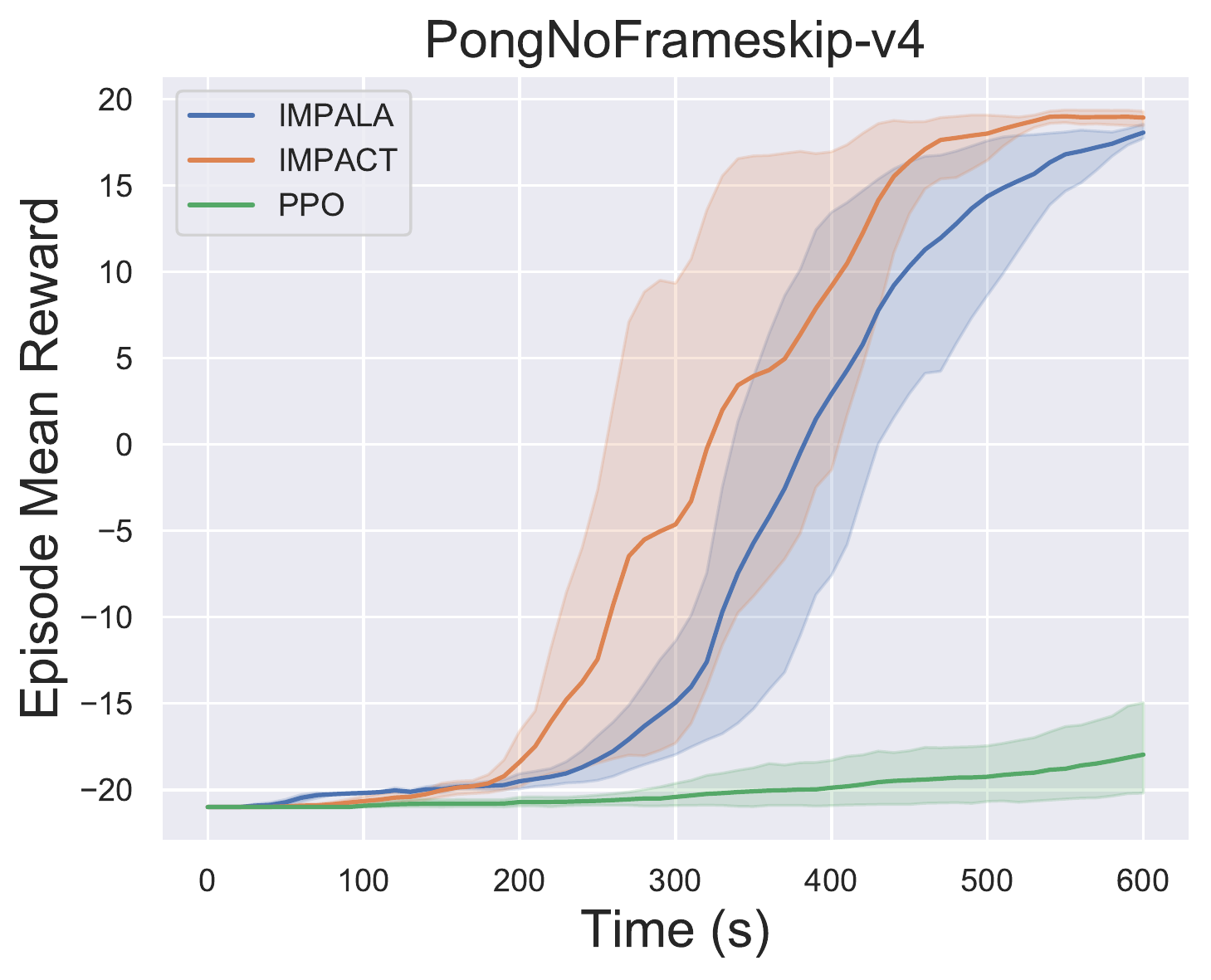}
        % \vspace*{-5mm}
        \subcaption{\small Time}
    \end{subfigure}
        % \vskip\baselineskip
    \begin{subfigure}{\textwidth}
        \centering
        \includegraphics[width=0.31\textwidth, height=0.15\textheight, clip]{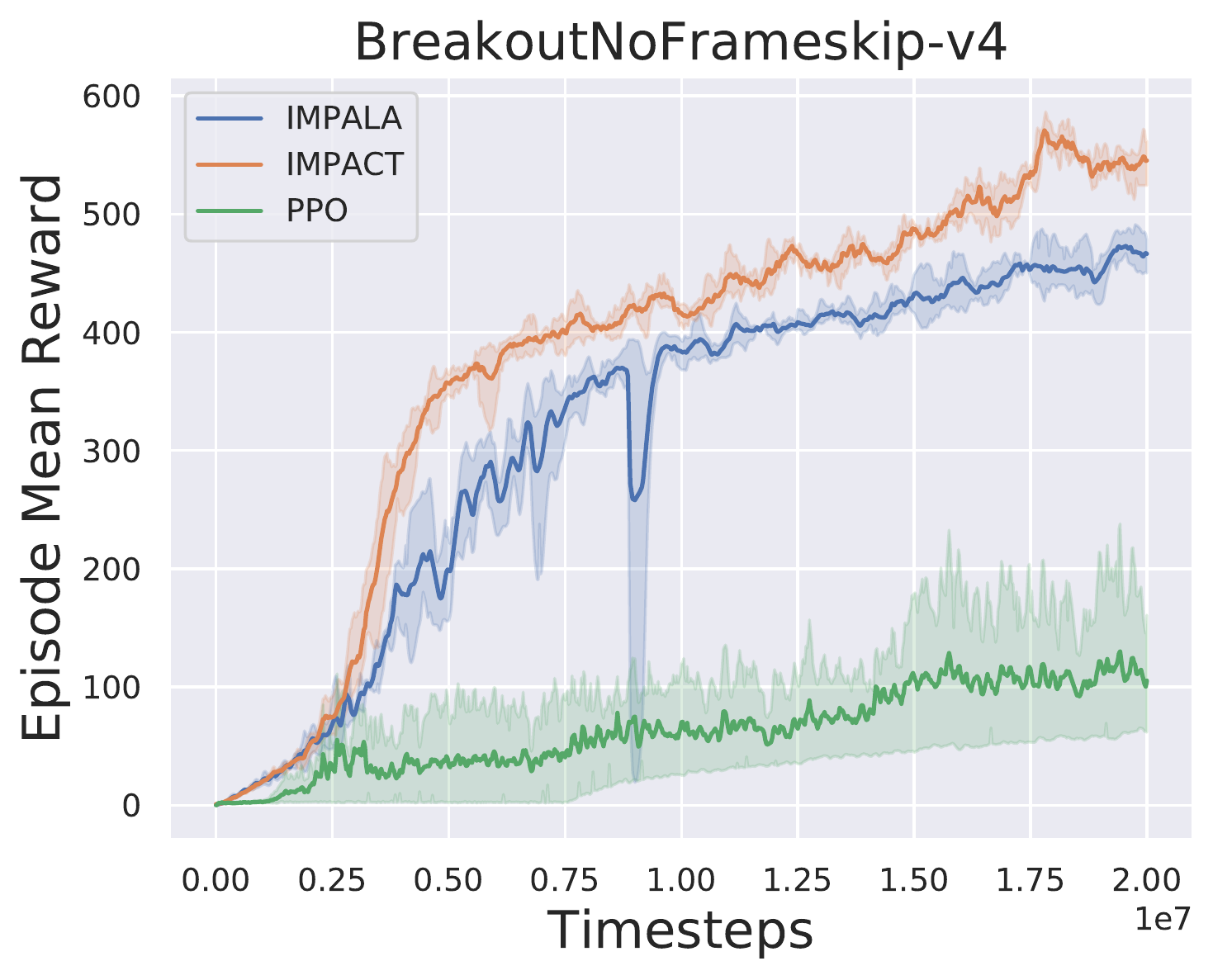}
        \hfill
        \includegraphics[width=0.31\textwidth, height=0.15\textheight, clip]{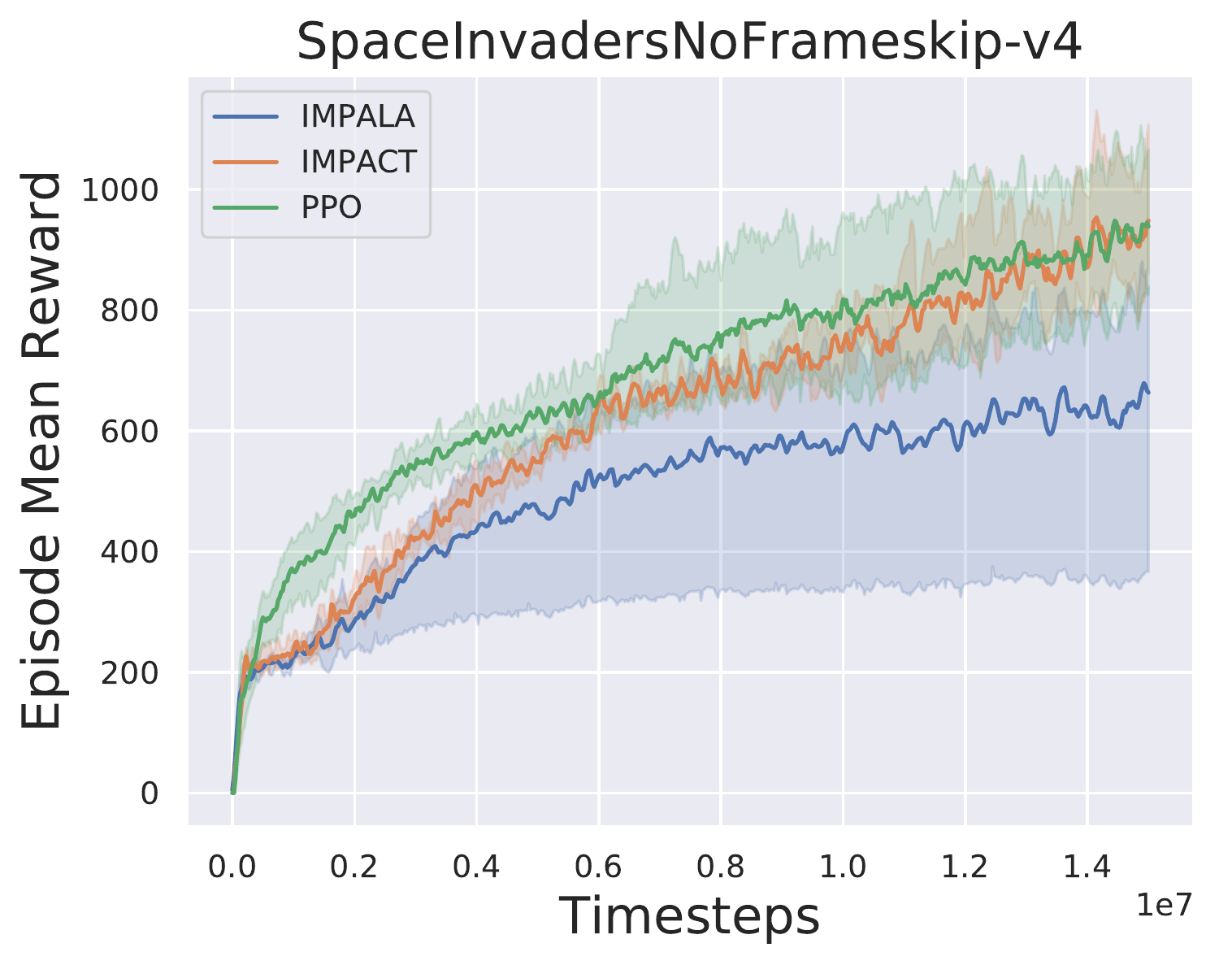}
        \hfill
        \includegraphics[width=0.31\textwidth, height=0.15\textheight, clip]{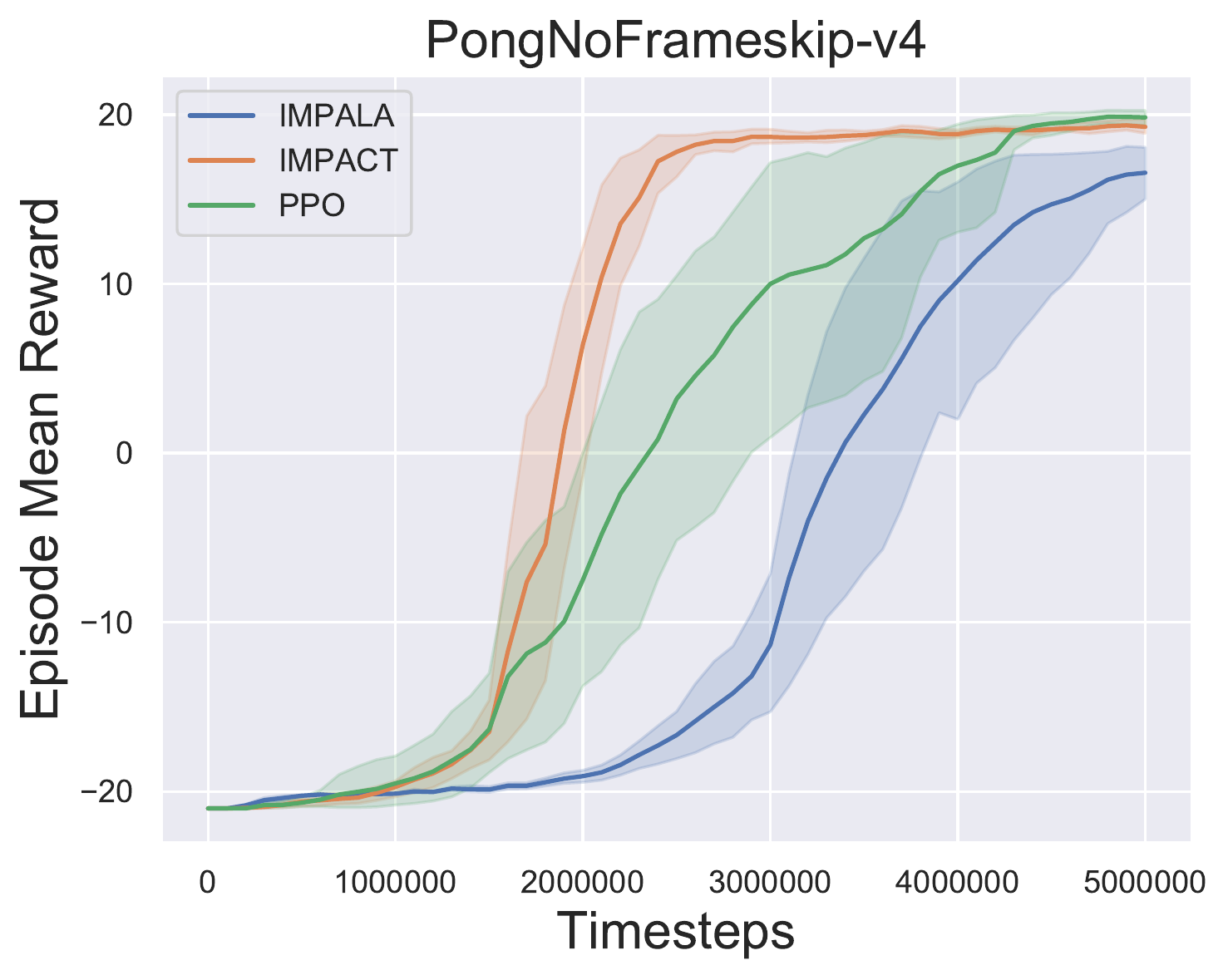}
        % \vspace*{-5mm}
        \subcaption{\small Timesteps}
    \end{subfigure}
    \caption{\agentname{} outperforms PPO and IMPALA in both real-time and sample efficiency for Discrete Control Domains: Breakout, SpaceInvaders, and Pong.}
    \captionsetup{belowskip=0pt}
	\label{fig:training_curves}
	\hspace{2mm}
	\vspace{-.5cm}
 \end{figure}

\subsection{IMPACT Scalability}
\begin{figure}[h]
    \centering
    \begin{subfigure}[t]{0.43\textwidth}
        \centering
        \includegraphics[width=\textwidth]{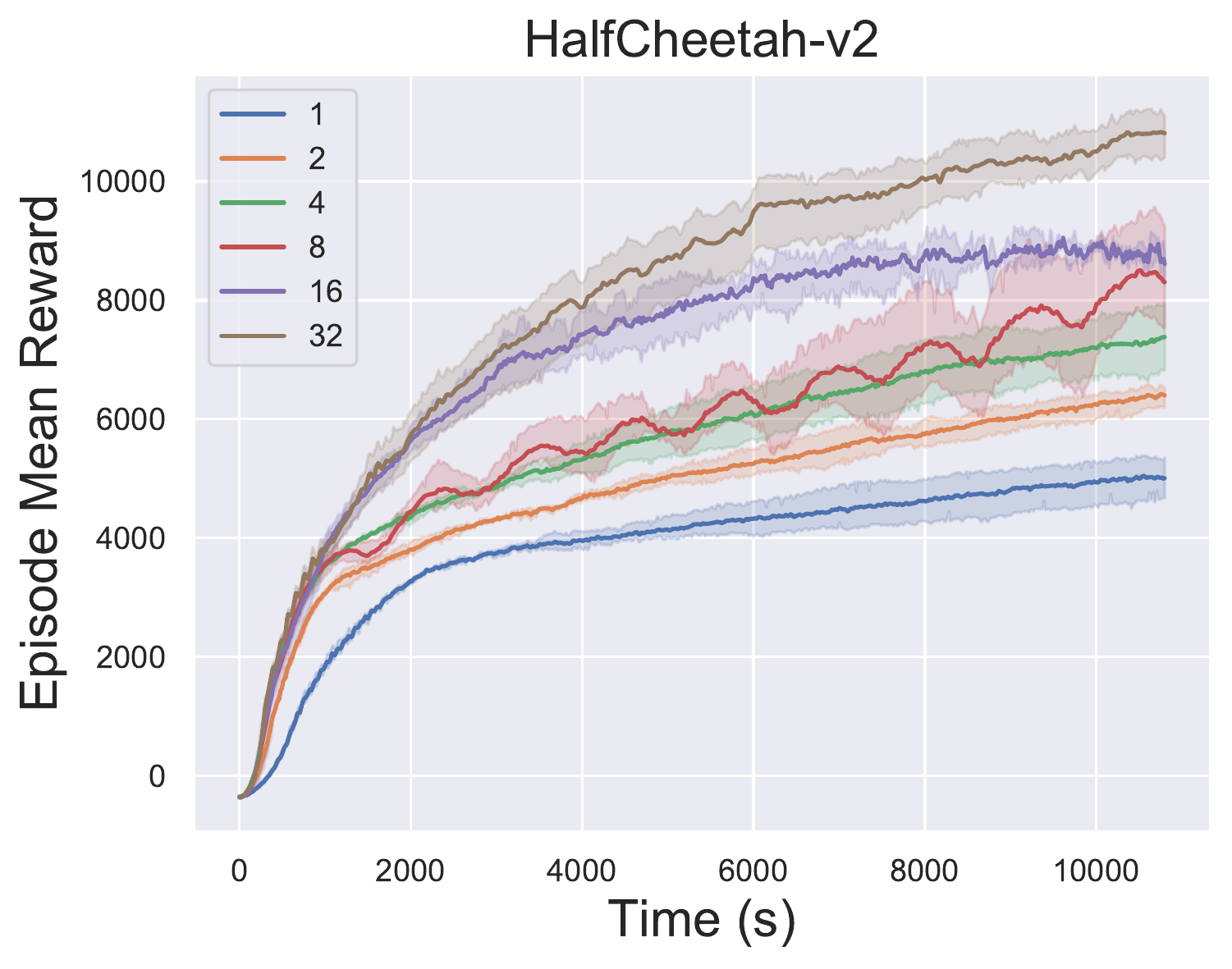}
        \caption{Continuous environment.}
        \label{fig:cont-env}
    \end{subfigure}
    % \hspace{20mm}
    \begin{subfigure}[t]{0.42\textwidth}
        \centering
        \includegraphics[width=\textwidth]{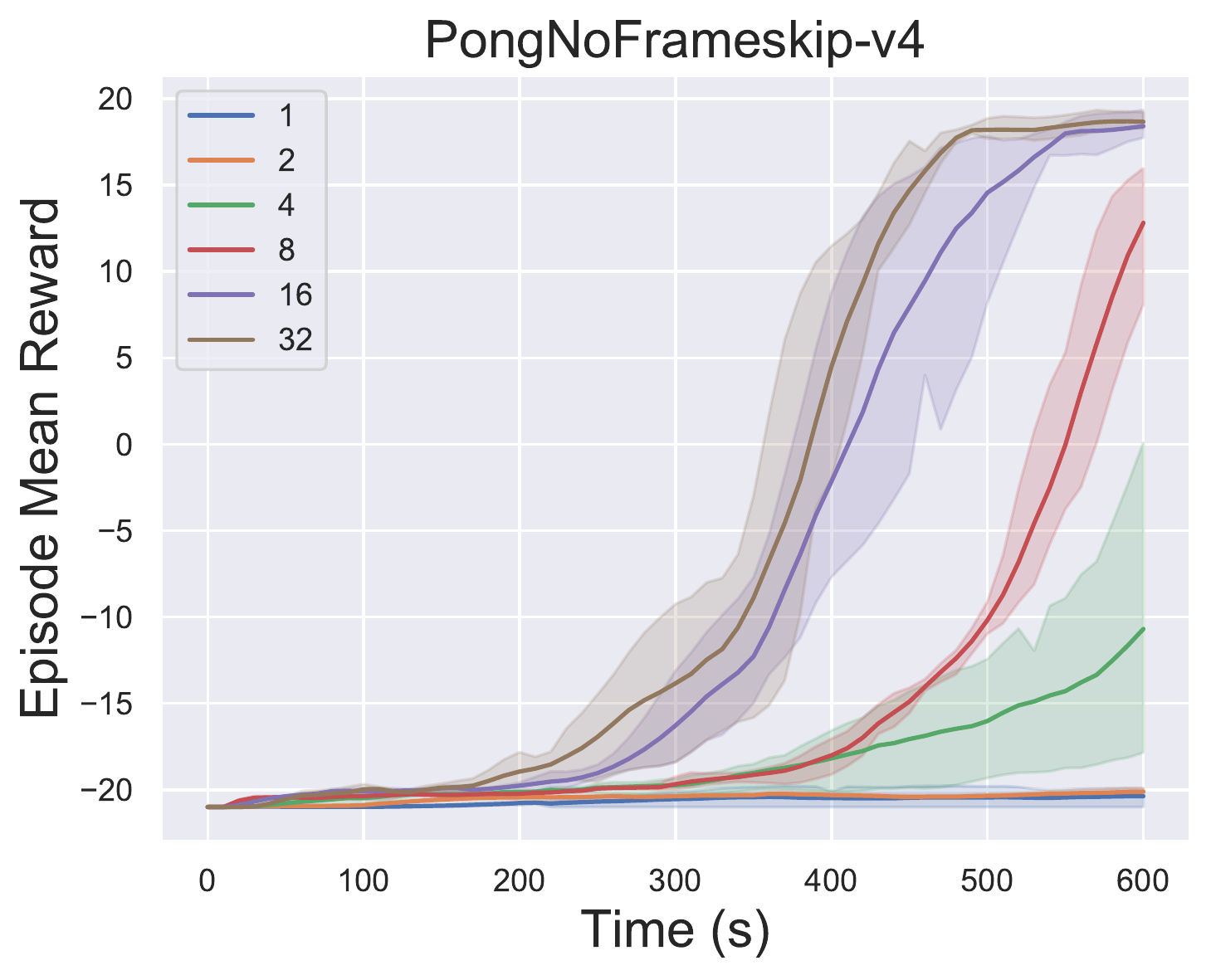}
        \caption{Discrete environment.}
        \label{fig:discre-env}
    \end{subfigure}
     \caption{Performance of IMPACT with respect to the number of workers in both continuous and discrete control tasks}
	\label{fig:scale}
	\hspace{-2mm}
	\vspace{-4mm}
\end{figure}

Figure \ref{fig:scale} shows how IMPACT's performance scales relative to the number of workers. More workers means increased sample throughput, which in turn increases training throughput (the rate that learner consumes batches). With the learner consuming more worker data per second, IMPACT can attain better performance in less time. However, as number of workers increases, observed increases in performance begin to decline.

% \begin{figure*}[H]

\section{Related Work}

\textbf{Distributed RL architectures} are often used to accelerate training. Gorila \citep{nair2015massively} and A3C \citep{mnih2016asynchronous} use workers to compute gradients to be sent to the learner. A2C \citep{mnih2016asynchronous} and IMPALA \citep{espeholt2018impala} send experience tuples to the learner. Distributed replay buffers, introduced in ACER \citep{wang2016sample} and Ape-X \citep{horgan2018distributed}, collect worker-collected experience and define an overarching heuristic for learner batch selection. IMPACT is the first to fully incorporate the sample-efficiency benefits of PPO in an asynchronous setting.

%\textbf{IMPALA and GA3C} (\cite{babaeizadeh2016reinforcement}) utilize GPUs for training separately from the sample collection process. By moving training off this logical flow path, the agent can significantly increase its sample throughput and decreases wall-clock training time. Despite fast training, IMPALA and A3C suffer from poor sample efficiency and perform sub-optimally on complex continuous control environments (\cite{tassa2018deepmind}). Off-policy experience contains stale samples which can be detrimental to the training process. 

\textbf{Surreal PPO} \citep{fan2018surreal} also studies training with PPO in the asynchronous setting, but do not consider adaptation of the surrogate objective nor IS-correction. Their use of a target network for broadcasting weights to workers is also entirely different from IMPACT's. Consequently, IMPACT is able to achieve better results in both real-time and sample efficiency.

%IMPALA employs V-trace for advantage estimation. \rliaw{are there any downsides to this? for the reader, this seems then like a solved problem}

\textbf{Off-policy methods}, including DDPG and QProp, utilize target networks to stabilize learning the Q function \citep{lillicrap2015continuous,gu2016q}. This use of a target network is related but different from IMPACT, which uses the network to define a stable trust region for the PPO surrogate objective. %Other agents, such as TRPO and PPO, constrain policy optimization within a trust region and attain high sample efficiency via mini-batch SGD, effectively replaying training data. (\cite{schulman2017proximal},\cite{schulman2015trust}). On the other hand, Soft Actor Critic intelligently employs a max-entropy framework and two target networks to maximize exploration while retaining training stability (\cite{haarnoja2018soft}). Most importantly, variance reduction methods lead to stable training. GAE-$\lambda$ and Retrace, estimate biased but low variance estimators for advantages and values (\cite{schulman2015high}, \cite{munos2016safe}). Alternatively, clipping importance-sampling ratios, such as PPO's surrogate loss and V-trace, mitigates off-policy estimates, thus decreases variance (\cite{espeholt2018impala}, \cite{schulman2017proximal}).

\section{Conclusion}

In conclusion, we introduce \agentname, which extends PPO with a stabilized surrogate objective for asynchronous optimization, enabling greater real-time performance without sacrificing timestep efficiency. We show the importance of the IMPACT objective to stable training, and show it can outperform tuned PPO and IMPALA baselines in both real-time and timestep metrics.

\newpage

\clearpage
\bibliography{iclr2020_conference}

\begin{thebibliography}{19}
\providecommand{\natexlab}[1]{#1}
\providecommand{\url}[1]{\texttt{#1}}
\expandafter\ifx\csname urlstyle\endcsname\relax
  \providecommand{\doi}[1]{doi: #1}\else
  \providecommand{\doi}{doi: \begingroup \urlstyle{rm}\Url}\fi

\bibitem[Achiam(2018)]{spinningup}
Joshua Achiam.
\newblock Openai {Spinning} {Up}.
\newblock \url{https://spinningup.openai.com/en/latest/spinningup/bench.html},
  November 2018.

\bibitem[Caspi et~al.(2017)Caspi, Leibovich, Novik, and Endrawis]{intelcoach}
Itai Caspi, Gal Leibovich, Gal Novik, and Shadi Endrawis.
\newblock Reinforcement {Learning} {Coach}, December 2017.
\newblock URL \url{https://doi.org/10.5281/zenodo.1134899}.

\bibitem[Espeholt et~al.(2018)Espeholt, Soyer, Munos, Simonyan, Mnih, Ward,
  Doron, Firoiu, Harley, Dunning, et~al.]{espeholt2018impala}
Lasse Espeholt, Hubert Soyer, Remi Munos, Karen Simonyan, Volodymir Mnih, Tom
  Ward, Yotam Doron, Vlad Firoiu, Tim Harley, Iain Dunning, et~al.
\newblock {IMPALA}: {Scalable} {Distributed} {Deep-RL} with {Importance}
  {Weighted} {Acto}-{Learner} {Architectures}.
\newblock \emph{arXiv preprint arXiv:1802.01561}, 2018.

\bibitem[Fan et~al.(2018)Fan, Zhu, Zhu, Liu, Zeng, Gupta, Creus-Costa,
  Savarese, and Fei-Fei]{fan2018surreal}
Linxi Fan, Yuke Zhu, Jiren Zhu, Zihua Liu, Orien Zeng, Anchit Gupta, Joan
  Creus-Costa, Silvio Savarese, and Li~Fei-Fei.
\newblock {SURREAL}: {Open}-{Source} {Reinforcement} {Learning} {Framework} and
  {Robot} {Manipulation} {Benchmark}.
\newblock In \emph{Conference on Robot Learning}, pp.\  767--782, 2018.

\bibitem[Gu et~al.(2016)Gu, Lillicrap, Ghahramani, Turner, and Levine]{gu2016q}
Shixiang Gu, Timothy Lillicrap, Zoubin Ghahramani, Richard~E Turner, and Sergey
  Levine.
\newblock {Q-Prop: Sample-Efficient Policy Gradient with An Off-Policy Critic
  }.
\newblock \emph{arXiv preprint arXiv:1611.02247}, 2016.

\bibitem[Han \& Sung(2017)Han and Sung]{han2017amber}
Seungyul Han and Youngchul Sung.
\newblock {AMBER: Adaptive Multi-Batch Experience Replay for Continuous Action
  Control}.
\newblock \emph{arXiv preprint arXiv:1710.04423}, 2017.

\bibitem[Han \& Sung(2019)Han and Sung]{han2019dimension}
Seungyul Han and Youngchul Sung.
\newblock {Dimension-Wise Importance Sampling Weight Clipping for
  Sample-Efficient Reinforcement Learning}.
\newblock \emph{arXiv preprint arXiv:1905.02363}, 2019.

\bibitem[Horgan et~al.(2018)Horgan, Quan, Budden, Barth-Maron, Hessel,
  Van~Hasselt, and Silver]{horgan2018distributed}
Dan Horgan, John Quan, David Budden, Gabriel Barth-Maron, Matteo Hessel, Hado
  Van~Hasselt, and David Silver.
\newblock {Distributed Prioritized Experience Replay}.
\newblock \emph{arXiv preprint arXiv:1803.00933}, 2018.

\bibitem[Jie \& Abbeel(2010)Jie and Abbeel]{jie2010connection}
Tang Jie and Pieter Abbeel.
\newblock {On a Connection between Importance Sampling and the Likelihood Ratio
  Policy Gradient}.
\newblock pp.\  1000--1008, 2010.

\bibitem[Liang et~al.(2018)Liang, Liaw, Nishihara, Moritz, Fox, Goldberg,
  Gonzalez, Jordan, and Stoica]{rllib}
Eric Liang, Richard Liaw, Robert Nishihara, Philipp Moritz, Roy Fox, Ken
  Goldberg, Joseph~E. Gonzalez, Michael~I. Jordan, and Ion Stoica.
\newblock {RLlib}: {Abstractions} for {Distributed} {Reinforcement} {Learning}.
\newblock In \emph{International Conference on Machine Learning ({ICML})},
  2018.

\bibitem[Lillicrap et~al.(2015)Lillicrap, Hunt, Pritzel, Heess, Erez, Tassa,
  Silver, and Wierstra]{lillicrap2015continuous}
Timothy~P Lillicrap, Jonathan~J Hunt, Alexander Pritzel, Nicolas Heess, Tom
  Erez, Yuval Tassa, David Silver, and Daan Wierstra.
\newblock Continuous {Control} with {Deep} {Reinforcement} {Learning}.
\newblock \emph{arXiv preprint arXiv:1509.02971}, 2015.

\bibitem[Mnih et~al.(2016)Mnih, Badia, Mirza, Graves, Lillicrap, Harley,
  Silver, and Kavukcuoglu]{mnih2016asynchronous}
Volodymyr Mnih, Adria~Puigdomenech Badia, Mehdi Mirza, Alex Graves, Timothy
  Lillicrap, Tim Harley, David Silver, and Koray Kavukcuoglu.
\newblock {Asynchronous Methods for Deep Reinforcement Learning}.
\newblock In \emph{International conference on machine learning}, pp.\
  1928--1937, 2016.

\bibitem[Nair et~al.(2015)Nair, Srinivasan, Blackwell, Alcicek, Fearon,
  De~Maria, Panneershelvam, Suleyman, Beattie, Petersen,
  et~al.]{nair2015massively}
Arun Nair, Praveen Srinivasan, Sam Blackwell, Cagdas Alcicek, Rory Fearon,
  Alessandro De~Maria, Vedavyas Panneershelvam, Mustafa Suleyman, Charles
  Beattie, Stig Petersen, et~al.
\newblock {Massively Parallel Methods for Deep Reinforcement Learning}.
\newblock \emph{arXiv preprint arXiv:1507.04296}, 2015.

\bibitem[Schulman et~al.(2015{\natexlab{a}})Schulman, Levine, Abbeel, Jordan,
  and Moritz]{schulman2015trust}
John Schulman, Sergey Levine, Pieter Abbeel, Michael Jordan, and Philipp
  Moritz.
\newblock {Trust Region Policy Optimization}.
\newblock In \emph{International conference on machine learning}, pp.\
  1889--1897, 2015{\natexlab{a}}.

\bibitem[Schulman et~al.(2015{\natexlab{b}})Schulman, Moritz, Levine, Jordan,
  and Abbeel]{schulman2015high}
John Schulman, Philipp Moritz, Sergey Levine, Michael Jordan, and Pieter
  Abbeel.
\newblock {High-Dimensional Continuous Control using Generalized Advantage
  Estimation}.
\newblock \emph{arXiv preprint arXiv:1506.02438}, 2015{\natexlab{b}}.

\bibitem[Schulman et~al.(2017)Schulman, Wolski, Dhariwal, Radford, and
  Klimov]{schulman2017proximal}
John Schulman, Filip Wolski, Prafulla Dhariwal, Alec Radford, and Oleg Klimov.
\newblock {Proximal Policy Optimization Algorithms}.
\newblock \emph{arXiv preprint arXiv:1707.06347}, 2017.

\bibitem[Sutton et~al.(2000)Sutton, McAllester, Singh, and
  Mansour]{sutton2000policy}
Richard~S Sutton, David~A McAllester, Satinder~P Singh, and Yishay Mansour.
\newblock {Policy Gradient Methods for Reinforcement Learning with Function
  Approximation}.
\newblock In \emph{Advances in neural information processing systems}, pp.\
  1057--1063, 2000.

\bibitem[Tassa et~al.(2018)Tassa, Doron, Muldal, Erez, Li, Casas, Budden,
  Abdolmaleki, Merel, Lefrancq, et~al.]{tassa2018deepmind}
Yuval Tassa, Yotam Doron, Alistair Muldal, Tom Erez, Yazhe Li, Diego de~Las
  Casas, David Budden, Abbas Abdolmaleki, Josh Merel, Andrew Lefrancq, et~al.
\newblock {Deepmind Control Suite}.
\newblock \emph{arXiv preprint arXiv:1801.00690}, 2018.

\bibitem[Wang et~al.(2016)Wang, Bapst, Heess, Mnih, Munos, Kavukcuoglu, and
  de~Freitas]{wang2016sample}
Ziyu Wang, Victor Bapst, Nicolas Heess, Volodymyr Mnih, Remi Munos, Koray
  Kavukcuoglu, and Nando de~Freitas.
\newblock {Sample Efficient Actor-Critic with Experience Replay}.
\newblock \emph{arXiv preprint arXiv:1611.01224}, 2016.

\end{thebibliography}
\bibliographystyle{iclr2020_conference}

\appendix

% \section{Proof of the Objective Function}
% \begin{eqnarray*}
% \operatorname{clip}(\frac{\pi_{\text{worker}_i}}{\pi_{\text{target}}}, 0, \beta) \frac{\pi_\theta}{\pi_{\text{worker}_i}} &=& 
% \min(\frac{\pi_{\text{worker}_i}}{\pi_{\text{target}}}, \beta) \frac{\pi_\theta}{\pi_{\text{worker}_i}}\\
%  &=& 
% \min(\frac{\pi_\theta }{\pi_{\text{target}}}, \beta\frac{\pi_\theta}{\pi_{\text{worker}_i}}) 
% \end{eqnarray*}
% or
% \begin{eqnarray*}
% \operatorname{clip}(\frac{\pi_{\text{worker}_i}}{\pi_{\text{target}}}, 0, \beta) \frac{\pi_\theta}{\pi_{\text{worker}_i}} &=& 
% \min(\frac{\pi_{\text{worker}_i}}{\pi_{\text{target}}}, \beta) \frac{\pi_\theta}{\pi_{\text{worker}_i}}\\
% &=& 
%  \frac{\pi_\theta}{\max(\frac{\pi_{\text{target}}}{\pi_{\text{worker}_i} }, \frac 1\beta) \pi_{\text{worker}_i} }\\
%  &=& 
%   \frac{\pi_\theta}{\max(\pi_{\text{target}} , \frac 1\beta \pi_{\text{worker}_i})  }\\
% \end{eqnarray*}

\section{Additional Experiments}

\begin{figure}[H]
    \centering
    \begin{subfigure}{\textwidth}
        \centering
        \includegraphics[width=0.31\textwidth, height=0.15\textheight, clip]{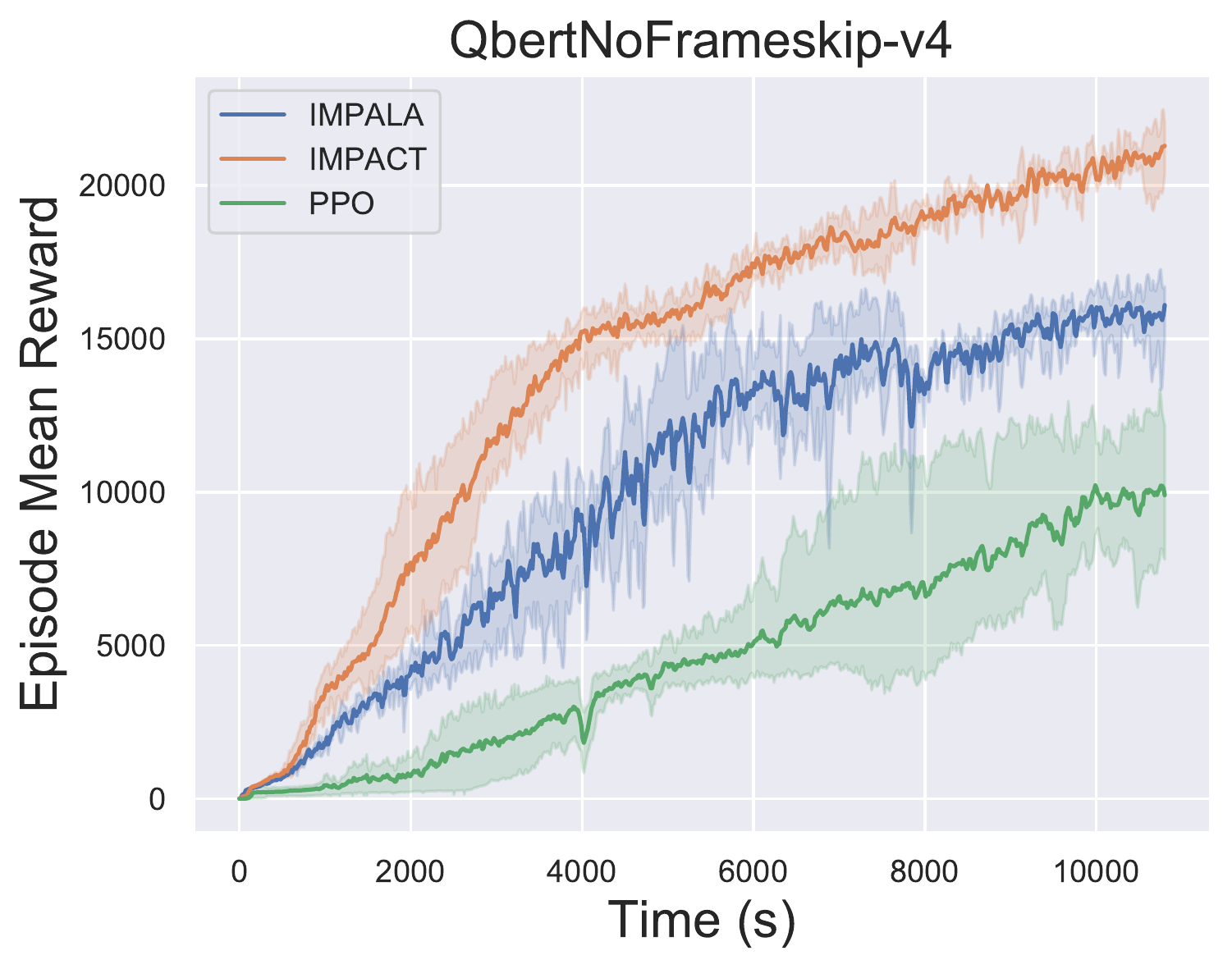}
        \hfill
        \includegraphics[width=0.31\textwidth, height=0.15\textheight, clip]{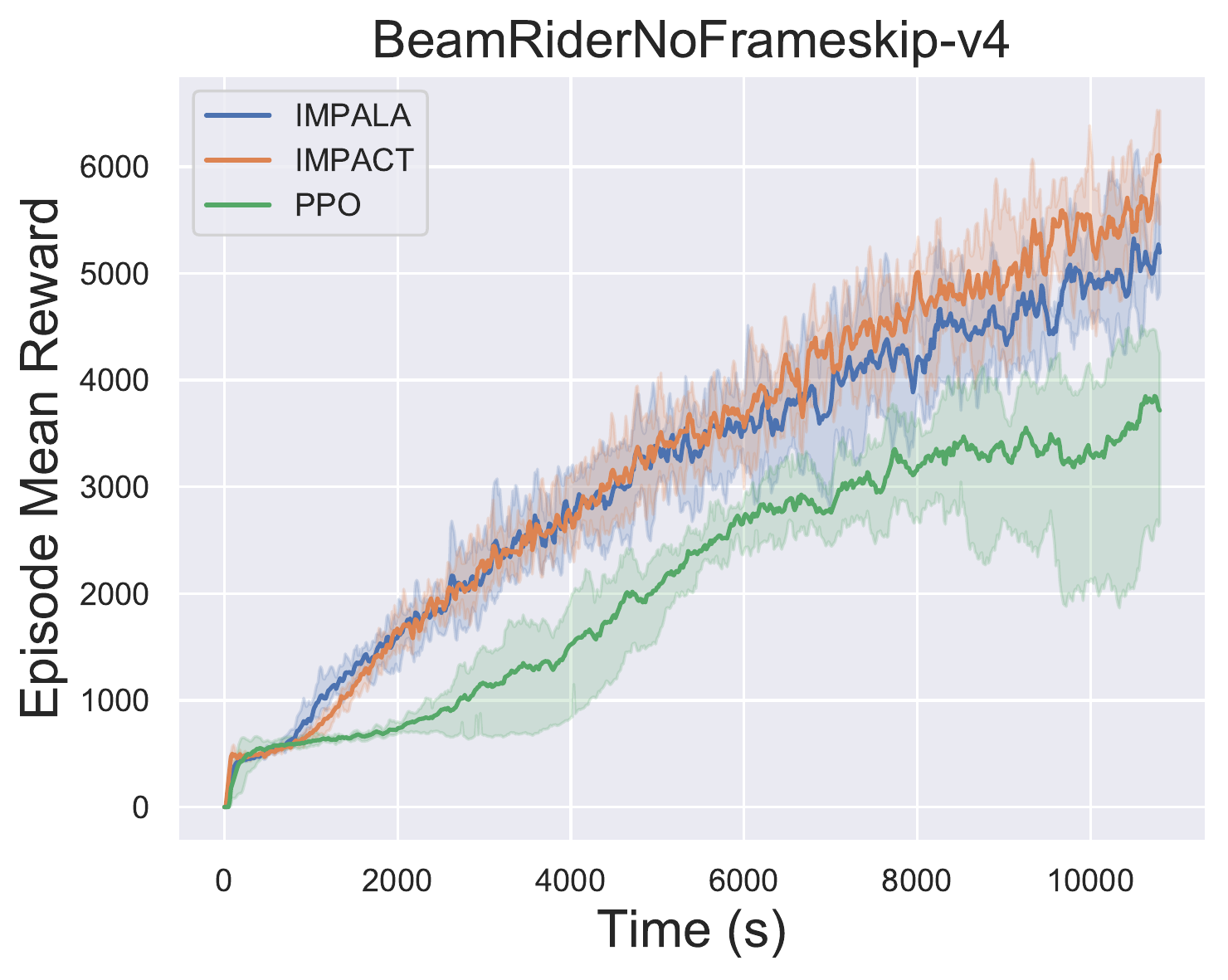}
        \hfill
        \includegraphics[width=0.31\textwidth, height=0.15\textheight, clip]{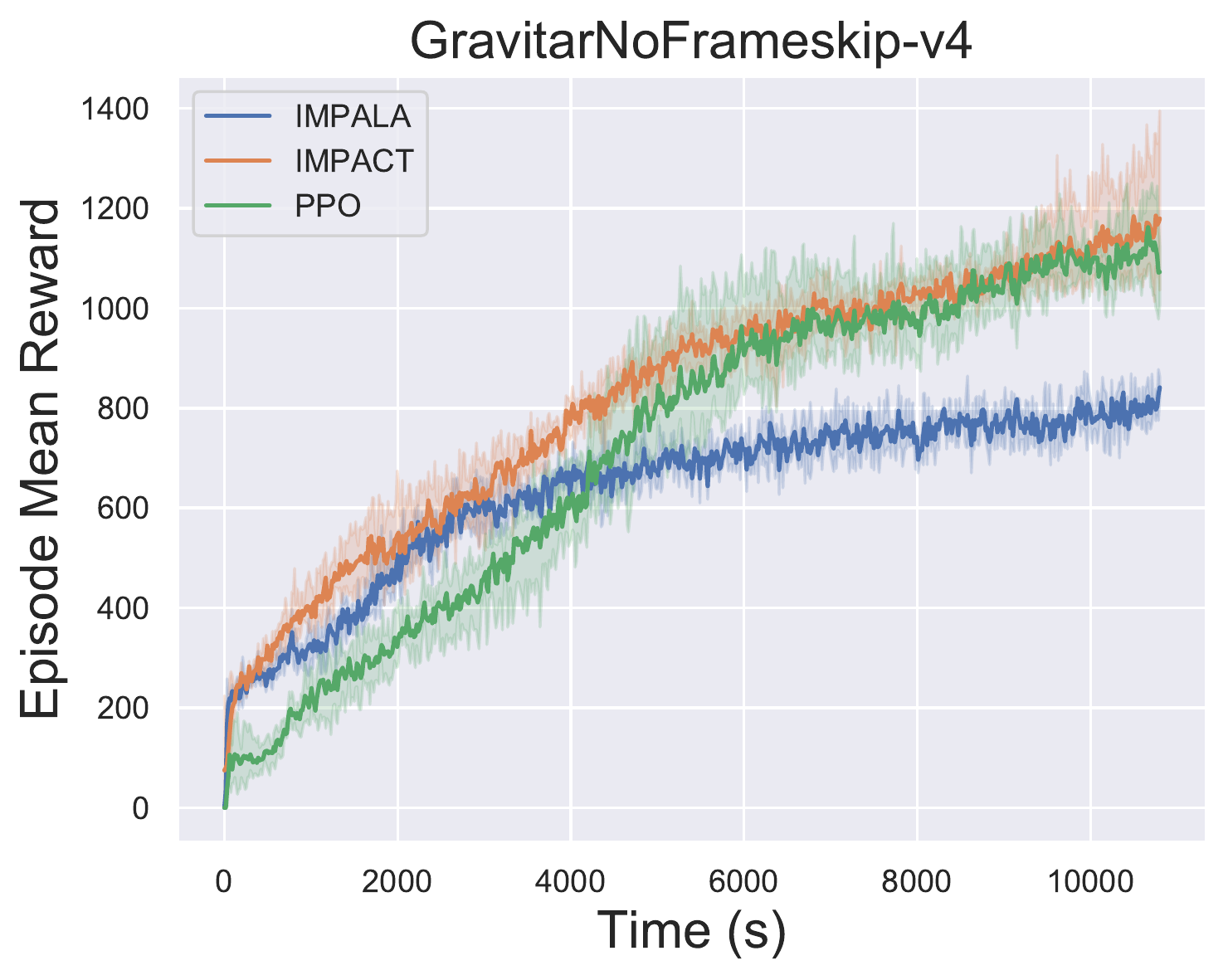}
        % \vspace*{-5mm}
        \caption{\small Time}
    \end{subfigure}
        % \vskip\baselineskip
    \begin{subfigure}{\textwidth}
        \centering
        \includegraphics[width=0.31\textwidth, height=0.15\textheight, clip]{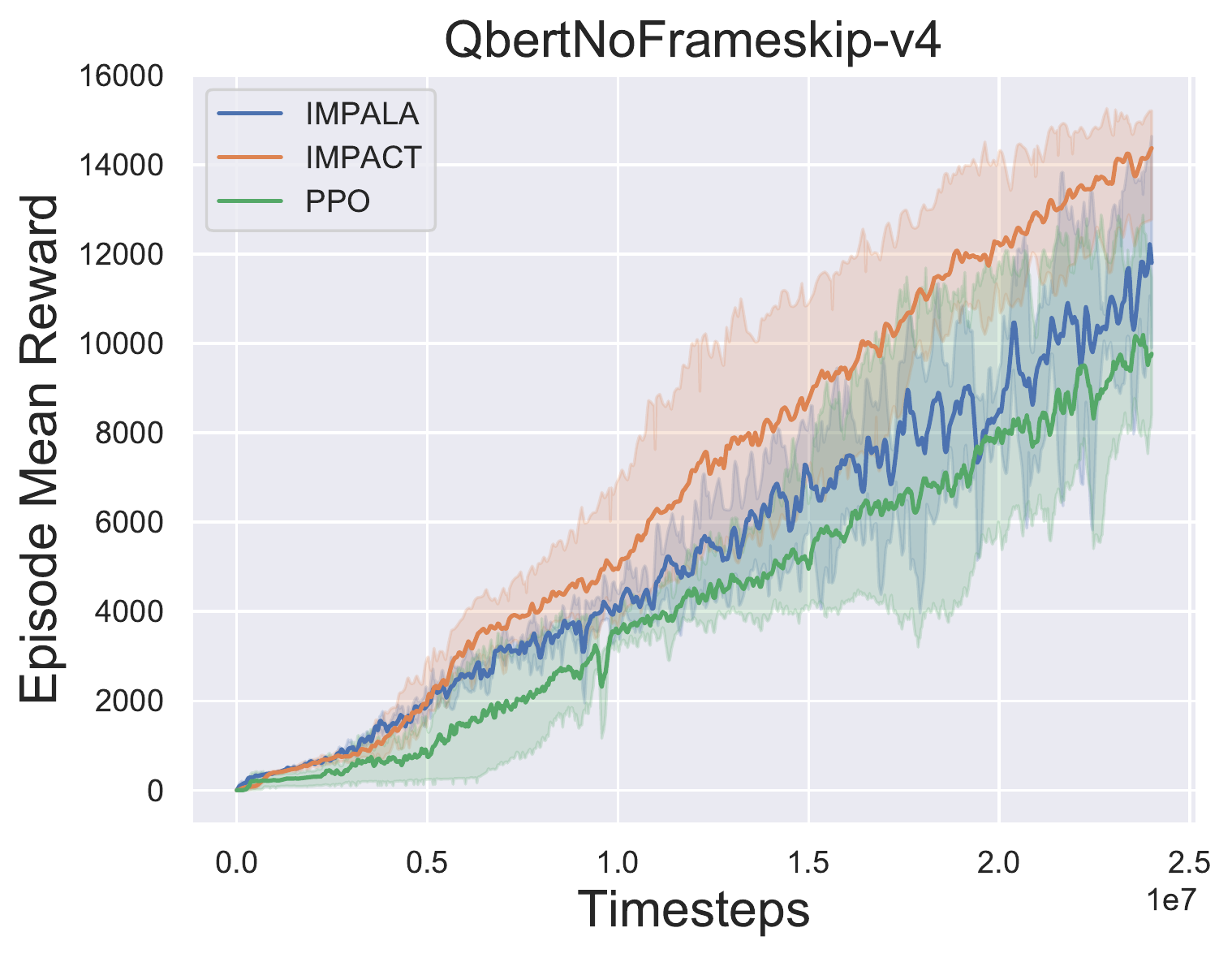}
        \hfill
        \includegraphics[width=0.31\textwidth, height=0.15\textheight, clip]{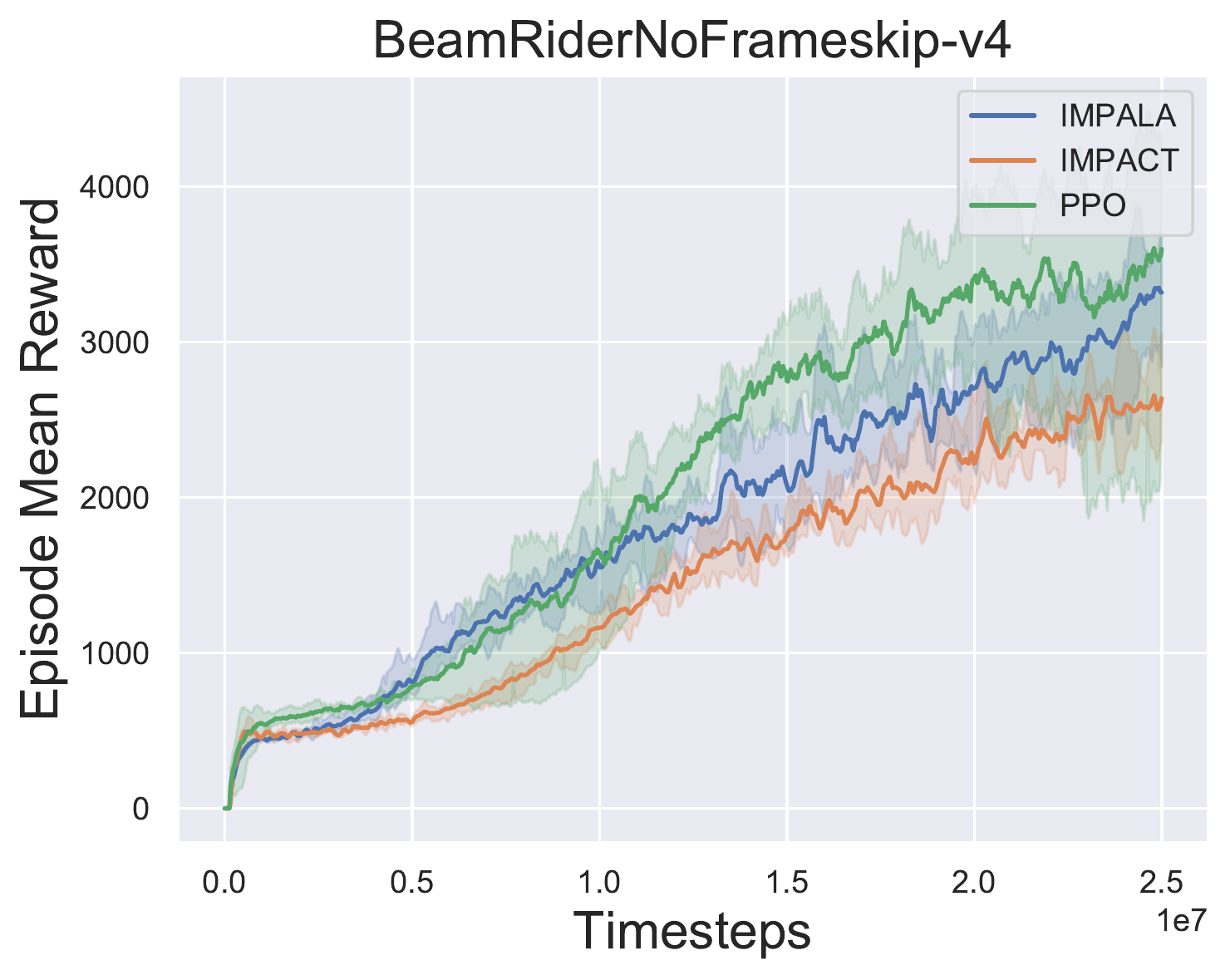}
        \hfill
        \includegraphics[width=0.31\textwidth, height=0.15\textheight,  clip]{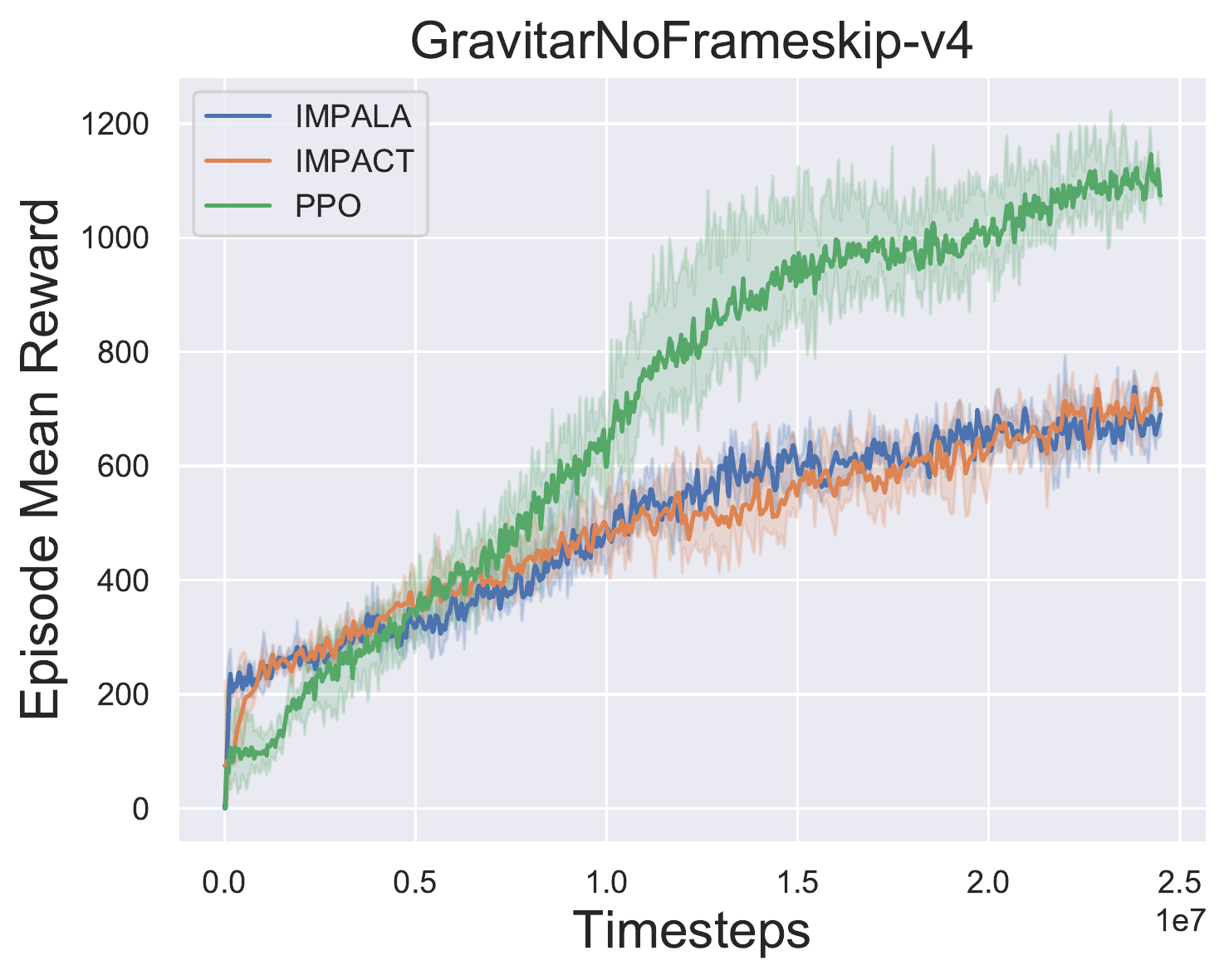}
        % \vspace*{-5mm}
        \caption{\small Timesteps}
    \end{subfigure}
    \caption{\agentname{}, PPO and IMPALA wallclock time and sample efficiency for Discrete Control Domains: Qbert, BeamRider, and Gravitar.}
	\label{fig:training_curves2}
	\hspace{2mm}
	\vspace{-.5cm}
 \end{figure}

% template of hyper
\section{Hyper parameters for All Environments}
\label{sec:exptparams}
\subsection{Discrete Environments}
\begin{table}[H]
\centering
\begin{tabular}{l|ccc}
\toprule
%\begin{sc}
{\sc Hyperparameters} & {\sc IMPACT} & {\sc IMPALA} & {\sc PPO} \\
%\end{sc}
\midrule
Clip Parameter & 0.3 & --- &0.1 \\
Entropy Coeff & 0.01 & 0.01 & 0.01 \\
Grad Clipping & 10.0 & 40.0 & --- \\
Discount ($\gamma$) & 0.99 & 0.99 & 0.99 \\
Lambda ($\lambda$)  & 0.995 & --- & 0.995 \\
Learning Rate & $1.0\cdot 10^{-4}$ & $1.0\cdot 10^{-4}$ & $5.0\cdot 10^{-5}$  \\
Minibatch Buffer Size (N) & 4 & --- & --- \\
Num SGD Iterations (K) & 2 & --- & 2  \\
Sample Batch Size & 50 & 50 & 100 \\
Train Batch Size & 500 & 500 &  5000 \\
SGD Minibatch Size & --- & --- & 500 \\
KL Coeff & 0.0 &  ---  & 0.5 \\
KL Target & 0.01 &  --- & 0.01 \\
Value Function Coeff & 1.0  & 0.5 & 1.0\\
Target-Worker Clipping ($\rho$) & 2.0  & --- & ---\\
\bottomrule
\end{tabular}

\caption{Hyperparameters for Discrete Environments.}
\label{table:discrete_control_settings}
\end{table}

\subsection{Continuous Environments}
\begin{table}[H]
\centering
\begin{tabular}{l|ccc}
\toprule
%\begin{sc}
{\sc Hyperparameters} & {\sc IMPACT} & {\sc IMPALA} & {\sc PPO} \\
%\end{sc}
\midrule
Clip Parameter & 0.4 & --- &0.3 \\
Entropy Coeff & 0.0 & 0.0 & 0.0 \\
Grad Clipping & 0.5 & 0.5 & --- \\
Discount ($\gamma$) & 0.995 & 0.995 & 0.99 \\
Lambda ($\lambda$)  & 0.995 & --- & 0.995 \\
Learning Rate & $3.0 \cdot 10^{-4}$ &$1.5\cdot10^{-5}$& $3.0 \cdot 10^{-4}$  \\
Minibatch Buffer Size (N) & 16 & --- & --- \\
Num SGD Iterations\protect\footnotemark (K) & 20 & --- & 20  \\
Sample Batch Size & 1024 & 1024 & 1024 \\
Train Batch Size & 32768 & 32768 &  163840\\
SGD Minibatch Size & --- & --- & 32768 \\
KL Coeff & 1.0 &  ---  & 1.0 \\
KL Target & 0.04 &  --- & 0.01 \\
Value Function Coeff\protect\footnotemark & 1.0  & 0.5 & 1.0\\
Target-Worker Clipping ($\rho$) & 2.0  & --- & ---\\
\bottomrule
\end{tabular}

\caption{Hyperparameters for Continuous Control Environments}
\label{table:continuous_control_settings}
\end{table}

\addtocounter{footnote}{-1}

\footnotetext{For HalfCheetah-v2, IMPACT and PPO Num SGD Iterations (K) is 32.}
\stepcounter{footnote}
\footnotetext{For HalfCheetah-v2, IMPACT Value Function Coeff is 0.5.}
\stepcounter{footnote}
\footnotetext{IMPALA was difficult to finetune due to unstable runs.}

\addtocounter{footnote}{-1}

\subsection{Hyperparameter Budget}
Listed below was the grid search we used for each algorithm to obtain optimal hyperparameters. Optimal values were found via grid searching on each hyperparameter separately. We found that IMPACT's optimal hyperparameter values tend to hover close to either IMPALA's or PPO's, which greatly mitigated IMPACT's budget.
\subsubsection{Discrete Environment Search}
\begin{table}[H]
\centerline{
\begin{tabular}{l|ccc}
\toprule
%\begin{sc}
{\sc Hyperparameters} & {\sc IMPACT} & {\sc IMPALA} & {\sc PPO} \\
%\end{sc}
\midrule
Clip Parameter & [0.1, 0.2, 0.3] & --- & [0.1, 0.2, 0.3, 0.4] \\
Grad Clipping & [10, 20, 40] & [2.5, 5, 10, 20, 40, 80] & --- \\
Learning Rate ($10^{-4}$) & [0.5, 1.0, 3.0] & [0.1, 0.3, 0.5, 0.8, 1.0, 3.0, 5.0] & [0.5, 1.0, 3.0, 5.0, 8.0]  \\
Minibatch Buffer Size (N) & [2,4,8, 16] & --- & --- \\
Num SGD Iterations (K) & [1,2,4] & --- & [1,2,4,8]  \\
Train Batch Size & --- & --- & [1000, 2500, 5000, 10000]\\
Value Function Coeff & [0.5, 1.0, 2.0]  & [0.25, 0.5, 1.0, 2.0] & [0.25, 0.5, 1.0, 2.0]\\
\midrule
\midrule
\# of Runs & 19  & 17 & 21\\
\bottomrule
\end{tabular}}
\caption{Hyperparameter Search for Discrete Environments}
\label{table:discrete_control_budget}
\end{table}

\subsection{Continuous Environment Search}

\begin{table}[H]
\centerline{\begin{tabular}{l|ccc}
\toprule
%\begin{sc}
{\sc Hyperparameters} & {\sc IMPACT} & {\sc IMPALA} & {\sc PPO} \\
%\end{sc}
\midrule
Clip Parameter & [0.2, 0.3, 0.4] & --- & [0.1, 0.2, 0.3, 0.4]\\
Grad Clipping & [0.5, 1.0, 5.0] & [0.1, 0.25, 0.5, 1.0, 5.0, 10.0] & --- \\
Learning Rate ($10^{-4}$) & [1.0, 3.0, 5.0] & [0.1, 0.15, 0.3, 0.5, 0.8, 1.0, 3.0, 5.0] \protect\footnotemark & [1.0, 3.0, 5.0] \\
Minibatch Buffer Size (N) & [4,8,16] & --- & --- \\
Num SGD Iterations (K) & [20,26,32] & --- & [20,26,32]  \\
Train Batch Size & --- & --- & [65536, 98304, 131072, 163840] \\
KL Target & [0.01, 0.02, 0.04] &  --- & [0.01, 0.02, 0.04] \\
Value Function Coeff & [0.5, 1.0, 2.0]  & [0.5, 1.0, 2.0] & [0.5, 1.0, 2.0]\\
\midrule
\midrule
\# of Runs & 21  & 17 & 20\\
\bottomrule
\end{tabular}
}

\caption{Hyperparameter Search for Continuous Environments}
\label{table:continuous_control_budget}

\end{table}

% \begin{itemize}
%   \item IMPACT
%   \begin{itemize}
%   \item Clip Parameter - [0.2, 0.3, 0.4]
%   \item Learning Rate - [$1.0\cdot10^{-4}$, $3.0\cdot10^{-4}$, $8.0\cdot10^{-4}$]
%   \item Minibatch Buffer Size - [4,8,16]
%   \item Num SGD Iterations - [1,2,4]
%   \item Value Function Coeff - [0.5, 1.0, 2.0]
% %   \item num sgd iter
% \end{itemize}
%   \item IMPALA
%   \begin{itemize}
%   \item IMPACT
%   \item gradient clipping: [0.2, 0.5, 1.0]
%   \item value function loss coeff: [0.1, 0.5, 1.0, 2.0]
%   \item value function share layers: [False, True]
% \end{itemize}
%   \item PPO
% \begin{itemize}
%   \item IMPACT
%   \item PPO
%   \item Grad Clipping - 
%   \item KL divergence coefficient --- []
% \end{itemize}
% \end{itemize}
% \subsubsection{Continuous}

\section{IMPALA to IMPACT}
\begin{figure}[H]
% \centering\includegraphics[width=0.43\paperwidth,scale=0.31]{experiments/impala-impact-2.pdf}
\centering\includegraphics[width=0.63\paperwidth,scale=0.41]{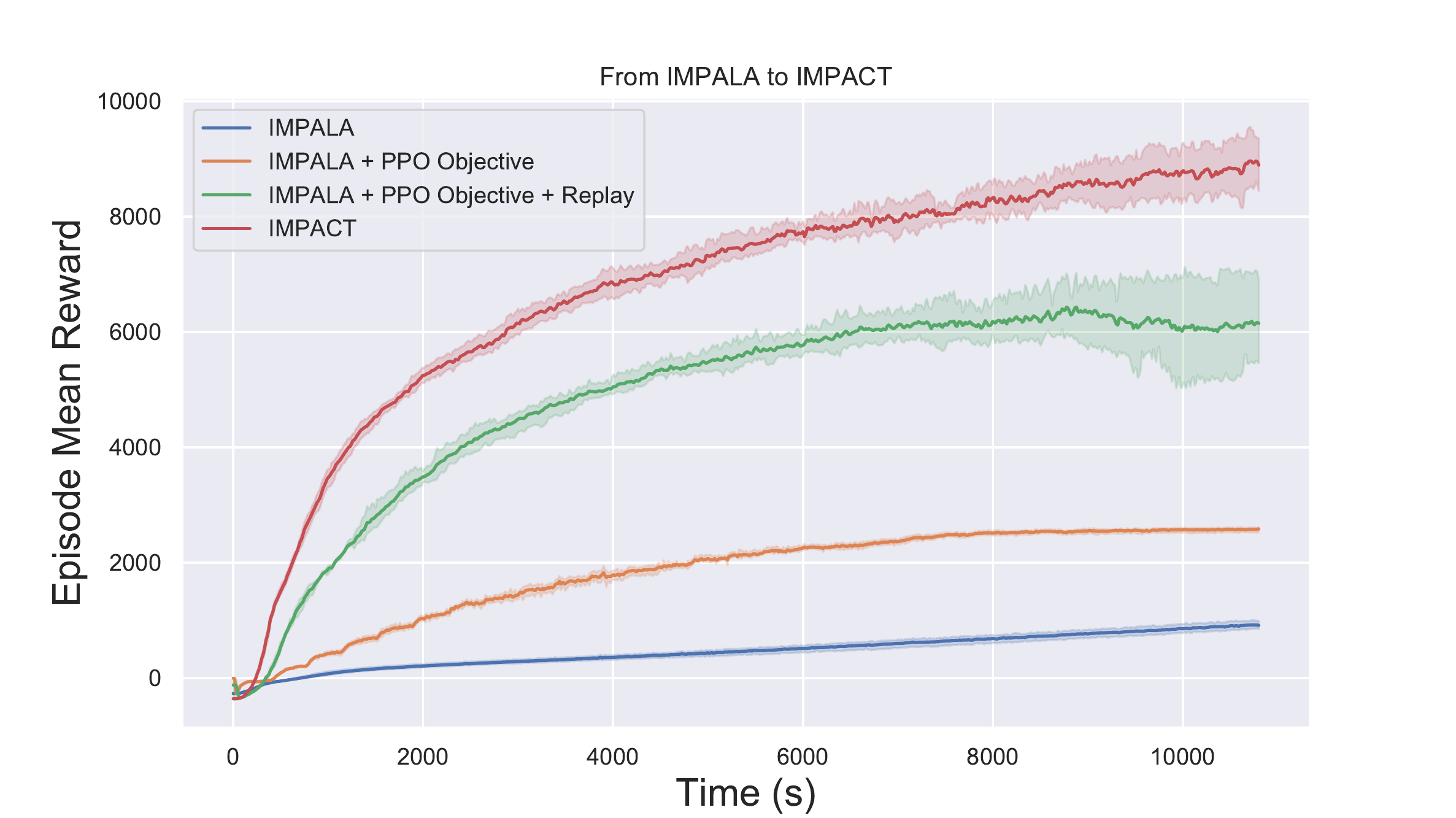}
% \centering\includegraphics[width=0.63\paperwidth,scale=0.41]{experiments/im2im2.png}
\caption{IMPALA to IMPACT: Incrementally Adding PPO Objective, Replay, and Target-Worker Clipping to IMPALA. The experiments are done on the HalfCheetah-v2 gym environment.}
\label{fig:impala2impact}
\end{figure}

In Figure \ref{fig:impala2impact}, we gradually add components to IMPALA until the agent is equivalent to IMPACT's. Starting from IMPALA, we gradually add PPO's objective function, circular replay buffer, and target-worker clipping. In particular, IMPALA with PPO's objective function and circular replay buffer is equivalent to an asynchronous-variant of PPO (APPO). APPO fails to perform as well as synchronous distributed PPO, since PPO is an on-policy algorithm.

\section{IMPALA in Continuous Environments}

In Figure \ref{fig:training_curves}, IMPALA performs substantially worse than other agents in continuous environments. We postulate that IMPALA suffers from low asymptotic performance here since its objective is an importance-sampled version of the Vanilla Policy Gradient (VPG) objective, which is known to suffer from high variance and large update-step sizes. We found that for VPG, higher learning rates encourage faster learning in the beginning but performance drops to negative return later in training. In Appendix \ref{sec:exptparams}, for IMPALA, we heavily tuned on the learning rate, finding that small learning rates stabilize learning at the cost of low asymptotic performance. Prior work also reveals the agents that use VPG fail to attain good performance in non-trivial continuous tasks \citep{spinningup}. Our results with IMPALA reaches similar performance compared to other VPG-based algorithms. The closest neighbor to IMPALA, A3C uses workers to compute gradients from the VPG objective to send to the learner thread. A3C performs well in InvertedPendulum yet flounders in continuous environments \citep{tassa2018deepmind}.

\begin{minipage}{\textwidth}
    \centering
    \begin{minipage}{0.48\textwidth}
        \centering
        \includegraphics[width=\textwidth]{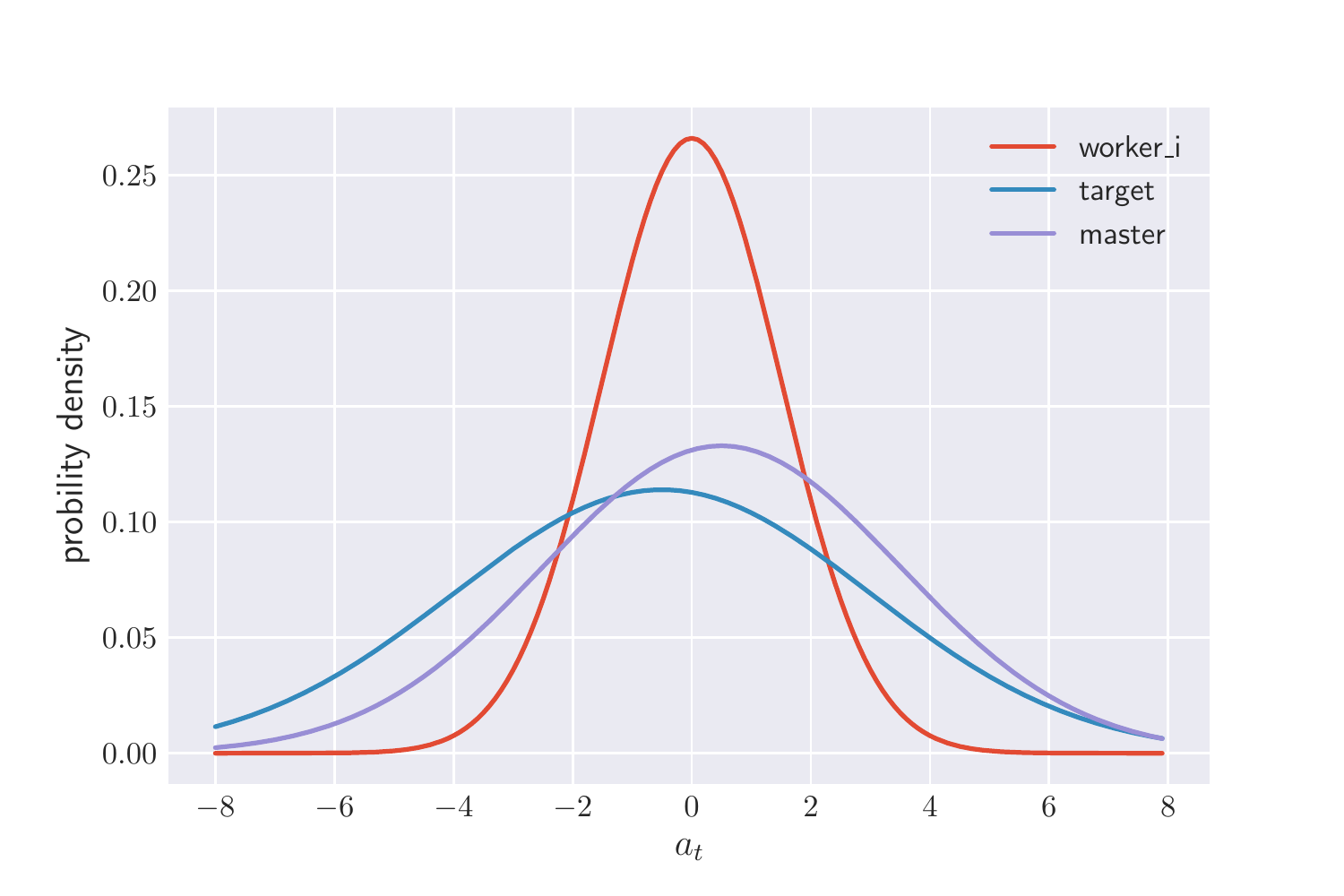}
    \captionof*{a}{Action Distributions}
    \end{minipage}
    \begin{minipage}{0.48\textwidth}
        \centering
        \includegraphics[width=\textwidth]{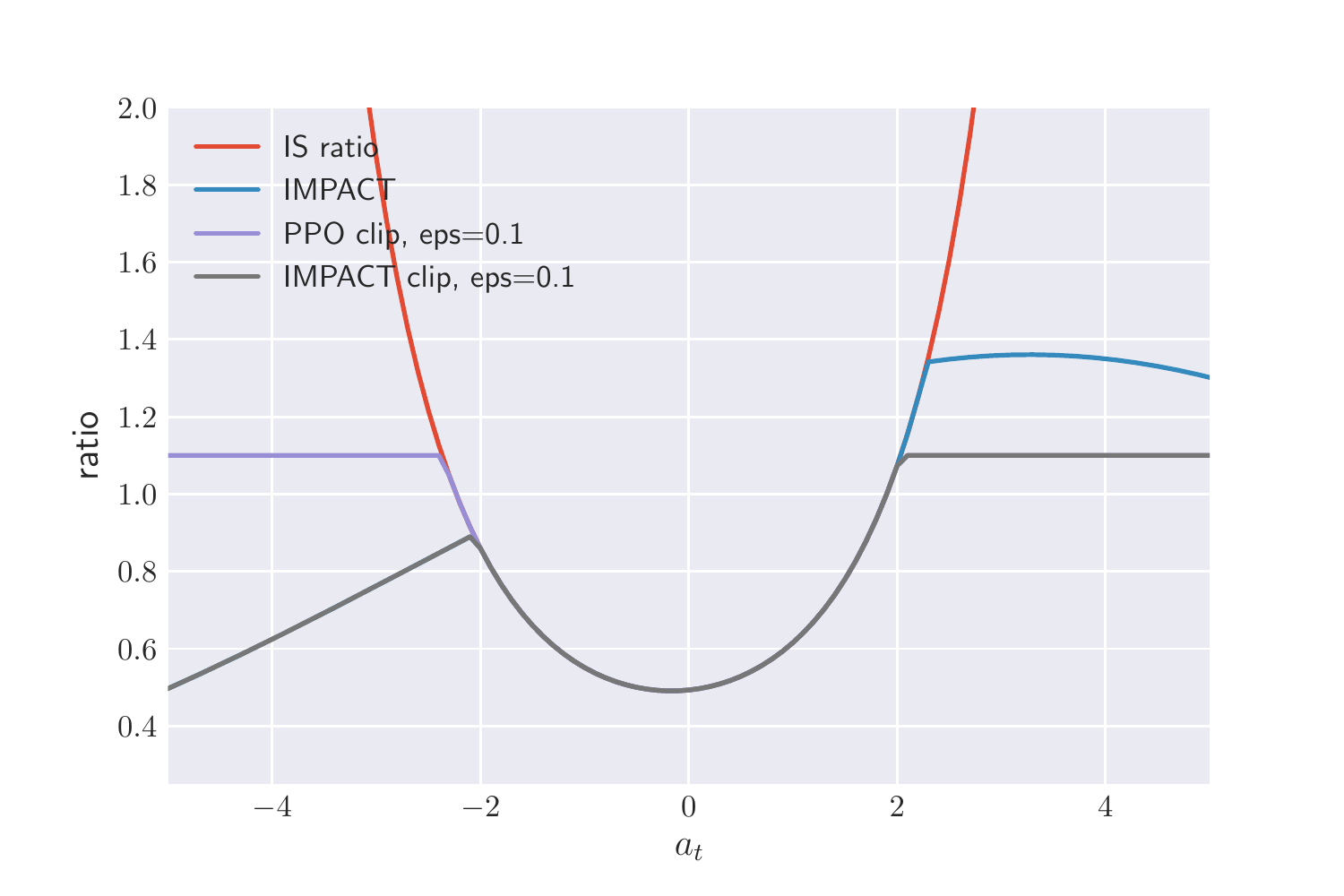}
    \captionof*{b}{Likelihood Ratios w.r.t Different Objectives}
    \end{minipage}
    \captionof{figure}{\small Likelihood ratio $r_t(\theta)$ for different objective functions, including PPO's. We assume a diagonal Gaussian policy for our policy. \textbf{Left}: Corresponding one dimensional action distributions for Worker i, Target, and Master Learner; \textbf{Right}: Ratio values graphed as a function of possible action values. IMPACT with PPO clipping is a lower bound of PPO. }
	\label{fig:intuition curves}
	\hspace{2mm}
 \end{minipage}

\section{The intuition of the objective}
The following ratios represent the objective functions for different ablation studies. In the plots (Figure \ref{fig:intuition curves}), we set the advantage function to be one, i.e. $\hat{A}_t=1.$

\begin{itemize}
\item IS ratio: $\frac{\pi_\theta}{\worker}\hat{A}_t$
\item \agentname \ target: $\min\left(\frac{\worker}{\target} , \rho\right)\frac{\pi_\theta}{\worker}\hat{A}_t$
\item PPO $\epsilon$-clip:  $\min\left( \frac{\pi_\theta}{\worker}\hat{A}_t, \text{clip}(\frac{\pi_\theta}{\worker}, 1-\epsilon, 1 + \epsilon)\hat{A}_t \right)$
\item \agentname{} target $\epsilon$-clip: $\min\left( \min\left(\frac{\worker}{\target} , \rho\right)\frac{\pi_\theta}{\worker}\hat{A}_t, \text{clip}\left(\min\left(\frac{\worker}{\target} , \rho\right)\frac{\pi_\theta}{\worker}, 1-\epsilon, 1 + \epsilon\right)\hat{A}_t \right)$
\end{itemize}

According to Figure \ref{fig:intuition curves}, IS ratio is large when $\worker$ assigns low probability. \agentname{} target $\epsilon$-clip is a lower bound of the PPO $\epsilon$-clip. In an distributed asynchronous setting, the trust region suffers from larger variance stemming from off-policy data. \agentname{} target $\epsilon$-clip ratio mitigates this by encouraging conservative and reasonable policy-gradient steps.

\end{document}